%% file: main.tex
\documentclass{article}

\PassOptionsToPackage{numbers, compress}{natbib}

\usepackage[dvipsnames]{xcolor}

\definecolor{myblue}{RGB}{31, 119, 180}
\definecolor{myorange}{RGB}{255, 127, 14}
\definecolor{mygreen}{RGB}{44, 160, 44}
\definecolor{myred}{RGB}{214, 39, 40}

\usepackage[final,main]{neurips_2026}

\usepackage[utf8]{inputenc} %
\usepackage[T1]{fontenc}    %
\usepackage[allcolors=NavyBlue,colorlinks=true,backref=page]{hyperref}       %
\usepackage{url}            %
\usepackage{booktabs}       %
\usepackage{amsfonts}       %
\usepackage{nicefrac}       %
\usepackage{microtype}      %
\usepackage{xcolor}         %
\usepackage{xspace}
\usepackage{subcaption}
\usepackage{array}

\usepackage{wrapfig}
\usepackage{tabularx}

\usepackage{algorithm}
\usepackage{algpseudocode}

\usepackage{cite}
\usepackage{comment}
\usepackage{prettyref}
\usepackage{amsfonts}
\usepackage{makecell}
\usepackage{adjustbox}
\usepackage{multicol}
\usepackage{multirow}
\usepackage{makecell}
\usepackage{caption}

\usepackage{amsmath}
\usepackage{amssymb}
\usepackage{amsmath,mathtools,thmtools}
\usepackage{enumitem}

\usepackage{thm-restate}

\usepackage[textsize=footnotesize]{todonotes}
\usepackage{xspace}

\usepackage[most]{tcolorbox}
\definecolor{colorcomment}{RGB}{160, 190, 210}%
\makeatletter
\algnewcommand{\LineComment}[1]{\Statex \hskip\ALG@thistlm \(\triangleright\) 
{\color{colorcomment}#1}}
\makeatother

\makeatletter
\algnewcommand{\IndentLineComment}[1]{\Statex \hskip\ALG@tlm \(\triangleright\) {\color{colorcomment}#1}}
\makeatother
\tcbset{
  tracebox/.style={
    enhanced,
    breakable,
    colback=gray!3,
    colframe=gray!45,
    boxrule=0.4pt,
    arc=1mm,
    left=4pt,
    right=4pt,
    top=4pt,
    bottom=4pt,
    fonttitle=\bfseries,
    coltitle=black,
    colbacktitle=gray!12,
    attach boxed title to top left={xshift=3mm,yshift=-2mm},
    boxed title style={
      colback=gray!12,
      colframe=gray!45,
      boxrule=0.4pt,
      arc=1mm
    }
  }
}
\usepackage{seqsplit}
\newcommand{\smiles}[1]{\texttt{\seqsplit{#1}}}

\input{macros}

\title{
Active-GRPO: Adaptive Imitation and Self-Improving Reasoning for Molecular Optimization
}

\author{Xuefeng Liu\textsuperscript{1}\thanks{Equal Contribution. Correspondence to: Xuefeng Liu <\href{mailto:xfl@stanford.edu}{xfl@stanford.edu}>, Mingxuan Cao <\href{mailto:caom@uchicago.edu}{mcao@uchicago.edu}>} ,~\textbf{Mingxuan Cao\textsuperscript{2$\ast$}},~\textbf{Qinan Huang\textsuperscript{3}},
~\textbf{Thomas Brettin\textsuperscript{5}},
~\textbf{Rick L. Stevens\textsuperscript{4,5}},
~\textbf{Le Cong\textsuperscript{1}} \\
\textsuperscript{1}School of Medicine, Stanford University\\
\textsuperscript{2}Data Science Institute, University of Chicago\\
\textsuperscript{3}Pritzker School of Molecular Engineering, University of Chicago\\
\textsuperscript{4}Department of Computer Science, University of Chicago\\
\textsuperscript{5}Argonne National Laboratory 
}

\begin{document}
\maketitle

\input{abstract}

\input{introduction}

\input{background_problemStatement}

\input{algorithm}

\input{experiments}

\input{conclusion}

\input{acknowledgement}

\bibliographystyle{plainnat}
\bibliography{reference}

\clearpage
\onecolumn

\appendix
\clearpage

\input{supp_implementation}
\input{related_works}

\input{supp_experiment}

\end{document}

%% file: macros.tex
\newcommand{\algname}{{Active-GRPO}\xspace}

\newif\iffinal
\finaltrue

\iffinal
    \newcommand{\fix}[1]{#1}
    \newcommand{\YC}[1]{}
    \newcommand{\YCinline}[1]{}
    \newcommand{\XL}[1]{}
    \newcommand{\XLinline}[1]{}
    \newcommand{\AC}[1]{}
    \newcommand{\ACinline}[1]{}
    \newcommand{\MW}[1]{}
    \newcommand{\MWinline}[1]{}
    \newcommand{\note}[1]{}
    \newcommand{\pref}[1]{}
    
\else
    \newcommand{\fix}[1]{{\color{red} #1}}
    \newcommand{\YC}[1]{\todo[fancyline,color=NavyBlue!40]{YC: #1}\xspace}
    \newcommand{\YCinline}[1]{\textcolor{NavyBlue}{[YC: #1]}}
    \newcommand{\XL}[1]{\todo[fancyline,color=green!40]{XL: #1}\xspace}
    \newcommand{\XLinline}[1]{\textcolor{NavyBlue}{[XL: #1]}}
    \newcommand{\AC}[1]{\todo[fancyline,color=purple!40]{AC: #1}\xspace}
    \newcommand{\ACinline}[1]{\textcolor{purple}{[AC: #1]}}
    \newcommand{\MW}[1]{\todo[fancyline,color=blue!40]{MW: #1}\xspace}
    \newcommand{\MWinline}[1]{\textcolor{green}{[MW: #1]}}
    \newcommand{\note}[1]{{\color{purple}[XL: #1]}}
    
    \newcommand{\pref}[1]{{\color{blue}(\ref{#1})}}
\fi

\let\hat\widehat

\RequirePackage{dsfont}

%% file: abstract.tex
\begin{abstract}

Scientific reasoning is an increasingly important capability of large language models, yet improving the robustness and efficiency of training such reasoning remains a key open challenge. We study this problem in instruction-based molecular optimization, where answer-only supervised fine-tuning (SFT) collapses multi-step reasoning and reinforcement learning with verifiable rewards (RLVR) suffers from sparse feedback. Reference-guided Policy Optimization (RePO) mitigates both by anchoring policy updates to dataset-provided references, but its effectiveness is tightly coupled to reference quality: weak or misaligned references impose a performance ceiling. To overcome this ceiling, we propose active reasoning, a paradigm in which the policy actively decides, on a per-instance basis, \emph{when} to imitate a reference and \emph{when} to reinforce its own discoveries, while continuously upgrading \emph{what} it imitates. We instantiate this paradigm as {Active Group Relative Policy Optimization} (\algname), realized through two coupled mechanisms: \emph{active imitate-reinforce} and \emph{active referencing}. The former performs imitation learning when the reference still outperforms the policy's own candidates, and shifts to self-improvement via reinforcement learning once the policy has generated molecules that surpass the reference. The latter continuously upgrades the reference itself by 
replacing it with the best policy-generated candidate discovered so far, progressively raising the imitation target and ensuring that reference guidance remains informative---rather than restrictive---throughout training. Across TOMG-Bench \textsc{MolOpt}, \algname improves average SR$\times$Sim from 0.0959 for GRPO and 0.1665 for RePO to 0.1773 under matched three-seed evaluation, with statistically significant gains on LogP, MR, and QED.

\end{abstract}

%% file: introduction.tex
\section{Introduction}\label{sec:intro}

Large language models (LLMs) have rapidly emerged as general-purpose reasoning engines, demonstrating strong performance on tasks that demand multi-step deliberation rather than surface pattern matching \citep{grattafiori2024llama, openai2023gpt}. Through advances in chain-of-thought prompting \citep{wei2022chain}, supervised fine-tuning (SFT) on reasoning traces \citep{muennighoff2025s1, zelikman2022star}, and reinforcement learning with verifiable rewards (RLVR) \citep{guo2025deepseek,lambert2024tulu}, modern LLMs can now solve competition-level mathematics \citep{cobbe2021training, hendrycks2021measuring}, write and debug complex code \citep{chen2021evaluating,jimenez2023swe}, and conduct structured analyses across diverse domains. This progress has motivated a growing line of work that brings LLM reasoning to bear on scientific discovery \citep{ai4science2023impact, taylor2022galactica}, where success often hinges on navigating combinatorially large hypothesis spaces under domain-specific constraints. From hypothesis generation and experimental design to candidate screening in chemistry, biology, and materials science \citep{boiko2023autonomous, jablonka2024leveraging, stokes2020deep}, LLMs are increasingly positioned not as passive question answerers but as active reasoners that propose, evaluate, and refine scientific artifacts. Yet making such reasoning \emph{robust} and \emph{sample-efficient} to train remains a central open challenge—particularly in scientific domains where outputs must satisfy strict, programmatically verifiable constraints.

Among these scientific reasoning tasks, instruction-based molecular optimization has emerged as a particularly demanding testbed \citep{li2024tomg, li2026reference}. Given a source molecule and a natural-language instruction specifying desired property changes—for example, improving aqueous solubility while preserving binding affinity—the model must propose a structurally similar yet property-improved candidate \citep{jin2018learning, jin2020hierarchical}. This task lies at the heart of drug discovery \citep{lipinski2004navigating, stokes2020deep}, agrochemical design\citep{djoumboufeunang2023cheminformatics}, and materials development \citep{boiko2023autonomous, sanchez2018inverse}, while imposing tightly coupled constraints. Outputs must be syntactically valid molecules \citep{weininger1988smiles}, retain a high degree of structural similarity to the input scaffold \citep{bajusz2015tanimoto, bemis1996properties}, achieve measurable improvements across one or more—often competing—property objectives \citep{brown2019guacamol, huang2021therapeutics}, and faithfully follow the user's instruction \citep{edwards2022translation, li2024empowering}. Unlike open-ended generation, molecular optimization therefore requires constrained, multi-objective reasoning over structured chemical objects, with every candidate verifiable against programmatic property predictors and similarity metrics \citep{gao2022sample, polykovskiy2020molecular}.
 
Existing training paradigms for this setting exhibit characteristic failure. Answer-only SFT \citep{ouyang2022training, taori2023stanford} forces the model to memorize input–output mappings without articulating chemical rationale, collapsing multi-step reasoning \citep{chu2025sft,lin2023unlocking} and limiting generalization to unseen instruction styles. RLVR \citep{guo2025deepseek,lambert2024tulu}, which optimizes directly against programmatic property checkers, in principle preserves reasoning, but in practice suffers from sparse feedback \citep{andrychowicz2017hindsight,riedmiller2018learning}: under tight similarity constraints, most sampled molecules fail validity or similarity gates and receive zero reward, starving the policy of learning signal. Reference-guided Policy Optimization (RePO) \citep{li2026repo} mitigates both pathologies by anchoring policy updates to dataset-provided reference molecules \citep{jin2018learning,li2024tomg}, blending imitation and reward-based learning to densify the training signal and inherit the stability benefits of demonstration-based learning \citep{hester2018deep, nair2018overcoming, peng2019advantage, rajeswaran2017learning}. However, RePO's effectiveness is tied to the static quality of its references. When references are weak, noisy, or misaligned with the instruction \citep{belkhale2023data, gao2024impact}, the imitation signal actively pulls the policy away from better solutions it might otherwise discover, creating a performance ceiling bounded by the dataset rather than by the policy's true capability.

To overcome this ceiling, we propose \textit{active reasoning}, a training paradigm where \emph{active} refers to actively deciding when to imitate a reference, when to reinforce its own discoveries, and what target to imitate; and \emph{reasoning} refers to the deliberative \texttt{<think>}...\texttt{</think><answer>}...\texttt{</answer>} generation. 
We instantiate this paradigm as \textbf{Active Group Relative Policy Optimization} (\algname), which couples active reasoning with two mechanisms: \emph{active imitate-reinforce} and \emph{active referencing}. The active imitate-reinforce mechanism performs imitation learning when the reference still outperforms the policy's own candidates, and shifts to self-improvement via reinforcement learning once the policy has generated molecules that surpass the reference. The active referencing mechanism continuously upgrades the reference itself by replacing it with the best policy-generated candidate discovered so far, progressively raising the imitation target as training proceeds. Together, these mechanisms ensure that reference guidance remains \emph{informative} rather than \emph{restrictive}, transitioning the policy from learning \emph{from} references to learning \emph{beyond} them. By construction, this makes reference guidance robust across the spectrum of reference quality.

\begin{figure*}[t]
    \centering
    \includegraphics[width=\textwidth]{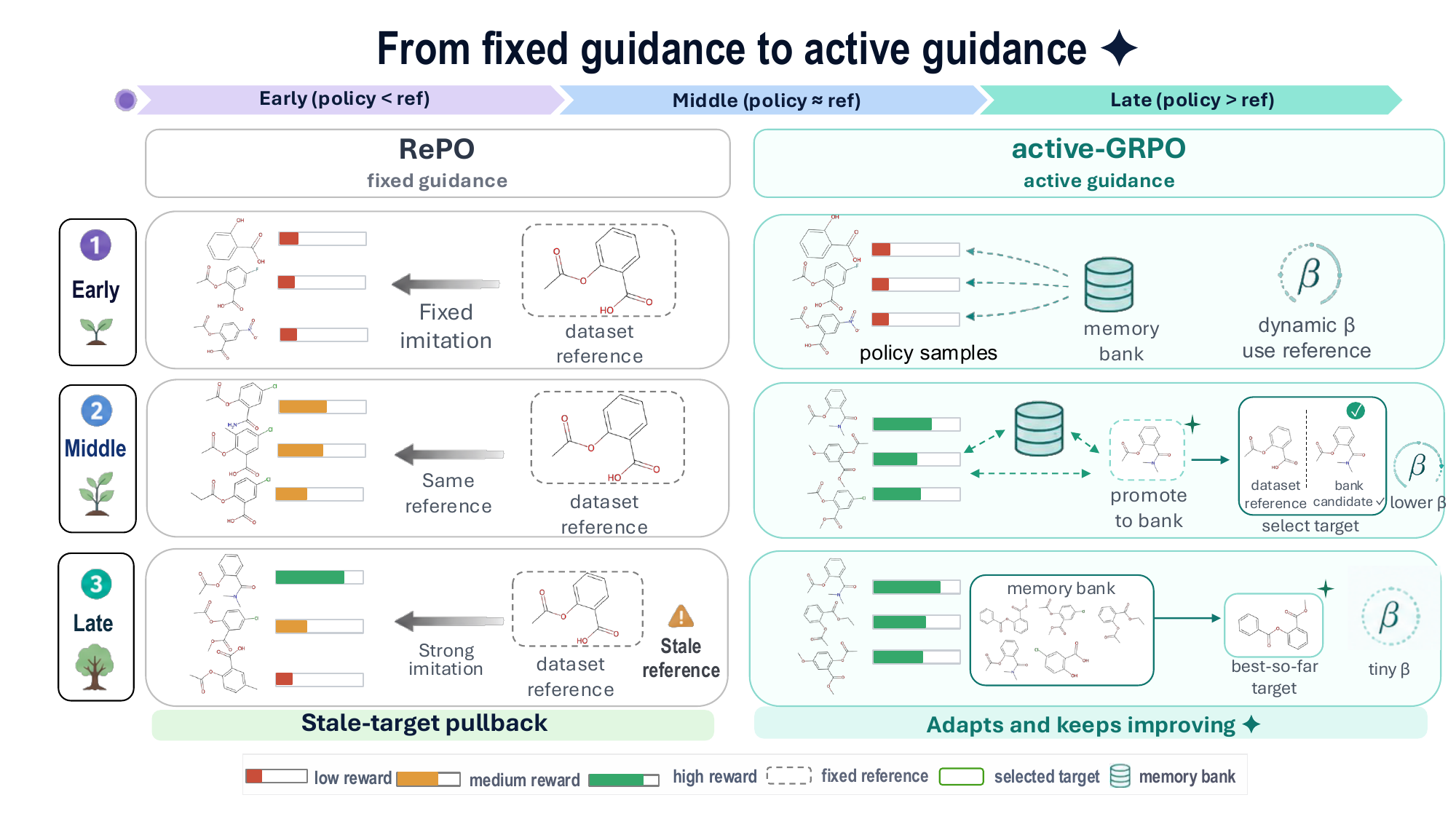}
    \caption{Conceptual motivation for \algname. RePO continues to imitate a fixed reference, which can become stale as the policy improves. \algname instead adapts both imitation strength and the guidance target, enabling a transition from reference imitation to active self-improvement.}
    \label{fig:fixed_to_active}
    \vspace{-0.5cm}
\end{figure*}
 
We evaluate \algname across a suite of molecular optimization benchmarks spanning diverse property objectives, instruction styles, and reference-quality regimes. Our contributions are threefold:
\begin{itemize}
    \item We identify and formally characterize the \emph{static-reference ceiling} in reference-guided policy optimization, showing that fixed dataset references can systematically mislead training when they fall below the policy's own capability.
    \item We introduce \emph{active reasoning} as a paradigm for reference-guided training, and instantiate it as \algname, which couples active imitate-reinforce and active referencing to make reference guidance robust to reference quality and self-improving over time.
    \item We show empirically that \algname consistently outperforms RePO and GRPO baselines, delivers more robust optimization across varying reference-quality regimes, and achieves a better balance across competing chemical objectives—establishing adaptive reference guidance as a principled path beyond the limits of static supervision.
\end{itemize}

%% file: background_problemStatement.tex
\vspace{-0.3cm}
\section{Preliminaries}\label{sec:Preliminaries}
\vspace{-0.3cm}

\subsection{{Problem Formulation: Instruction-Conditioned Molecular Generation}}
\label{sec:problem}
\vspace{-0.3cm}

We study {instruction-conditioned molecular generation with reference guidance}. The model receives a natural-language specification with task-dependent molecular context (typically an input molecule and optimization constraints), and produces a candidate molecule SMILES~\citep{weininger1988smiles} satisfying the specification. %
SMILES is a string representation of molecular graphs widely used in chemical language modeling and cheminformatics. Each training instance pairs a conditioning context $c_i$ with a dataset-provided reference molecule $m_{\mathrm{ref},i}$ providing answer-level guidance during training.

\vspace{-0.4cm}
\paragraph{Data and prompts.}

Let $\mathcal{D}=\{z_i\}_{i=1}^N$ be a training set, where each instance is
$z_i=(c_i,m_{\mathrm{ref},i}).$
Here $c_i$ is a task-dependent conditioning context, and $m_{\mathrm{ref},i}$ is a dataset-provided reference molecule. The two play different roles: $c_i$ specifies the optimization problem and conditions reward evaluation, whereas $m_{\mathrm{ref},i}$ serves as an answer-level guidance target during training.

\paragraph{Reasoning-augmented generation.}
\vspace{-0.5cm}
The policy $\pi_\theta$ is trained to produce a structured output interleaves a reasoning trace with a final answer: $o=\texttt{<think>}\;\tau\;\texttt{</think>}\;\texttt{<answer>}\;\hat m\;\texttt{</answer>}.$
Here $\tau$ is a free-form natural-language reasoning trace and $\hat m$ is the candidate molecule SMILES.
 This format follows recent reasoning-trained LLMs~\citep{guo2025deepseek} and is well suited to molecular optimization: the trace gives the model space to identify editable substructures, weigh modifications, and check constraints before committing to a final molecule. We do not supervise $\tau$ directly; only the final answer span carries explicit answer-level guidance (Section~\ref{sec:grpo_repo}).

\paragraph{Reward.}
\vspace{-0.5cm}
We assume a verifiable reward $R(\widehat{m}; c)$ defined on (candidate, context) pairs, with invalid or constraint-violating molecules receiving zero reward. In \textsc{MolOpt}, $R$ combines the requested property improvement with structural preservation; exact components are task-dependent. We also define the \emph{reference reward} $v_{\text{ref}}(z_i) = R(m_{\text{ref},i}; c_i)$, used as a per-instance baseline and as the anchor against which the policy's best candidates are compared in our method.

\vspace{-0.4cm}
\subsection{
{GRPO-based Reasoning Optimization}
}
\label{sec:grpo_repo}

\vspace{-0.4cm}
\paragraph{{GRPO.}}
{
Group Relative Policy Optimization (GRPO)~\citep{shao2024deepseekmath} is an actor-only variant of PPO~\citep{schulman2017proximal} that replaces the learned value baseline with within-group reward normalization.} {For each prompt $x_i$, GRPO samples $G$ rollouts $\{o_{i,j}\}_{j=1}^G$ from the old policy $\pi_{\theta_{\mathrm{old}}}$, extracts candidate molecules $\{\hat m_{i,j}\}_{j=1}^G$, and evaluates rewards}
$r_{i,j}=R(\hat m_{i,j};c_i),\bar r_i=\frac{1}{G}\sum_{j=1}^G r_{i,j}.$
{GRPO then forms the group-normalized advantage}
$\hat A_{i,j}=\frac{r_{i,j}-\bar r_i}{\sigma_{r,i}+\varepsilon}$,
where $\sigma_{r,i}$ is the within-group reward standard deviation.
{At the objective level, GRPO can be written as}
\begin{equation}
\begin{aligned}
\mathcal{J}_{\mathrm{GRPO}}(\theta)
&=
\mathbb{E}_{
\substack{
x_i\sim \mathcal{D},\\
\{o_{i,j}\}_{j=1}^G \sim \pi_{\theta_{\mathrm{old}}}(\cdot\mid x_i)
}}
\Bigg[
\frac{1}{G}\sum_{j=1}^G \frac{1}{|o_{i,j}|}
\sum_{t=1}^{|o_{i,j}|}
\Big(
\\
&\qquad\qquad
\min\!\Big[
\rho_{i,j,t}(\theta)\hat A_{i,j},
\,
\mathrm{clip}\!\bigl(\rho_{i,j,t}(\theta),1-\epsilon,1+\epsilon\bigr)\hat A_{i,j}
\Big]
-
\beta_{\mathrm{KL}}\,D^{\mathrm{KL}}_{i,j,t}
\Big)
\Bigg],
\end{aligned}
\end{equation}
{where}
$
\rho_{i,j,t}(\theta)
=
\frac{\pi_{\theta}(o_{i,j,t}\mid x_i,o_{i,j,<t})}
{\pi_{\theta_{\mathrm{old}}}(o_{i,j,t}\mid x_i,o_{i,j,<t})}.
$
{In implementation, we optimize the negative empirical counterpart of this objective. For clarity, we denote the resulting minibatch loss by}
\[
\mathcal{L}_{\mathrm{RL}}
=
-\frac{1}{\sum_{i,j,t} m^{\mathrm{cmp}}_{i,j,t}}
\sum_{i,j,t}
\Big[
\rho_{i,j,t}\hat A_{i,j}
-
\beta_{\mathrm{KL}}D^{\mathrm{KL}}_{i,j,t}
\Big]m^{\mathrm{cmp}}_{i,j,t},
\]
where $m^{\mathrm{cmp}}_{i,j,t}$ is the completion mask, $\rho_{i,j,t}$ is the per-token policy ratio, and $D^{\mathrm{KL}}_{i,j,t}$ is the per-token KL penalty against the frozen reference policy $\pi_{\mathrm{ref}}$.
{Because GRPO learns from within-prompt relative reward differences, its signal can weaken when all sampled rollouts for a prompt receive similar rewards. This motivates adding answer-level guidance in reference-guided molecular optimization.}

\paragraph{{GRPO with reference-molecule guidance.}}
RePO~\citep{li2026repo} {augments GRPO with an answer-level imitation loss that pulls the policy toward the dataset reference molecule, adding supervised signal to relative reward optimization.} Given a target molecule $m$, RePO defines
\[
\mathcal{L}_{\mathrm{guide}}^{(i)}(m)
=
-
\frac{
\sum_t \log \pi_\theta(m_t\mid x_i,m_{<t})\,m^{\mathrm{ans}}_{i,t}
}{
\sum_t m^{\mathrm{ans}}_{i,t}+\varepsilon
},
\]
where $m^{\mathrm{ans}}_{i,t}$ masks only the answer span; the reasoning trace $\tau$ is not supervised. In vanilla RePO, the guidance target is fixed to the dataset reference, $m_i^*=m_{\mathrm{ref},i}$, and the objective is
\[
\mathcal{L}_{\mathrm{RePO}}
=
\mathcal{L}_{\mathrm{RL}}
+
\frac{1}{B}\sum_{i=1}^B \mathcal{L}_{\mathrm{guide}}^{(i)}(m_{\mathrm{ref},i}).
\]

{\paragraph{Limitation: The Static-Reference Ceiling.}
RePO's design tacitly assumes that the dataset reference is consistently a useful target. This assumption breaks down in two practically common regimes. \emph{(i) Reference saturation:} once the policy starts generating molecules whose reward matches or exceeds $v_{\mathrm{ref}}(z_i)$, continuing to imitate $m_{\mathrm{ref},i}$ pulls the policy back toward a strictly worse target. \emph{(ii) Weak references:} when the dataset reference is itself far from optimal --- for example, when references are automatically curated or noisy --- guidance toward $m_{\mathrm{ref},i}$ caps achievable performance well below what the policy could otherwise reach. In both regimes, two design choices that fixed-reference guidance cannot make become first-class degrees of freedom: 
{\emph{when} to imitate at all}, and \emph{what} to imitate. Our method (Section~\ref{sec:algorithm}) makes both choices adaptive and per-instance.}

%% file: algorithm.tex
\section{Algorithm: \algname}
\label{sec:algorithm}

We now instantiate the active reasoning paradigm — letting the policy decide, per instance, when to imitate a reference and when to reinforce its own discoveries, while continuously upgrading what it imitates — as Active Group Relative Policy Optimization (\algname). \algname augments the RePO objective with two coupled mechanisms:
\begin{itemize}[noitemsep, wide,labelwidth=0pt, labelindent=0pt]
    \item \textbf{Active imitate-reinforce} decides \textit{when} to imitate via a smooth, context-dependent guidance weight that compares the policy's current best samples against the reference.
    \item \textbf{Active referencing} decides \textit{what} to imitate via a per-instance memory bank that promotes policy-generated candidates once they outperform the dataset reference.
\end{itemize}

\begin{figure*}[t]
    \centering
    \includegraphics[width=\textwidth]{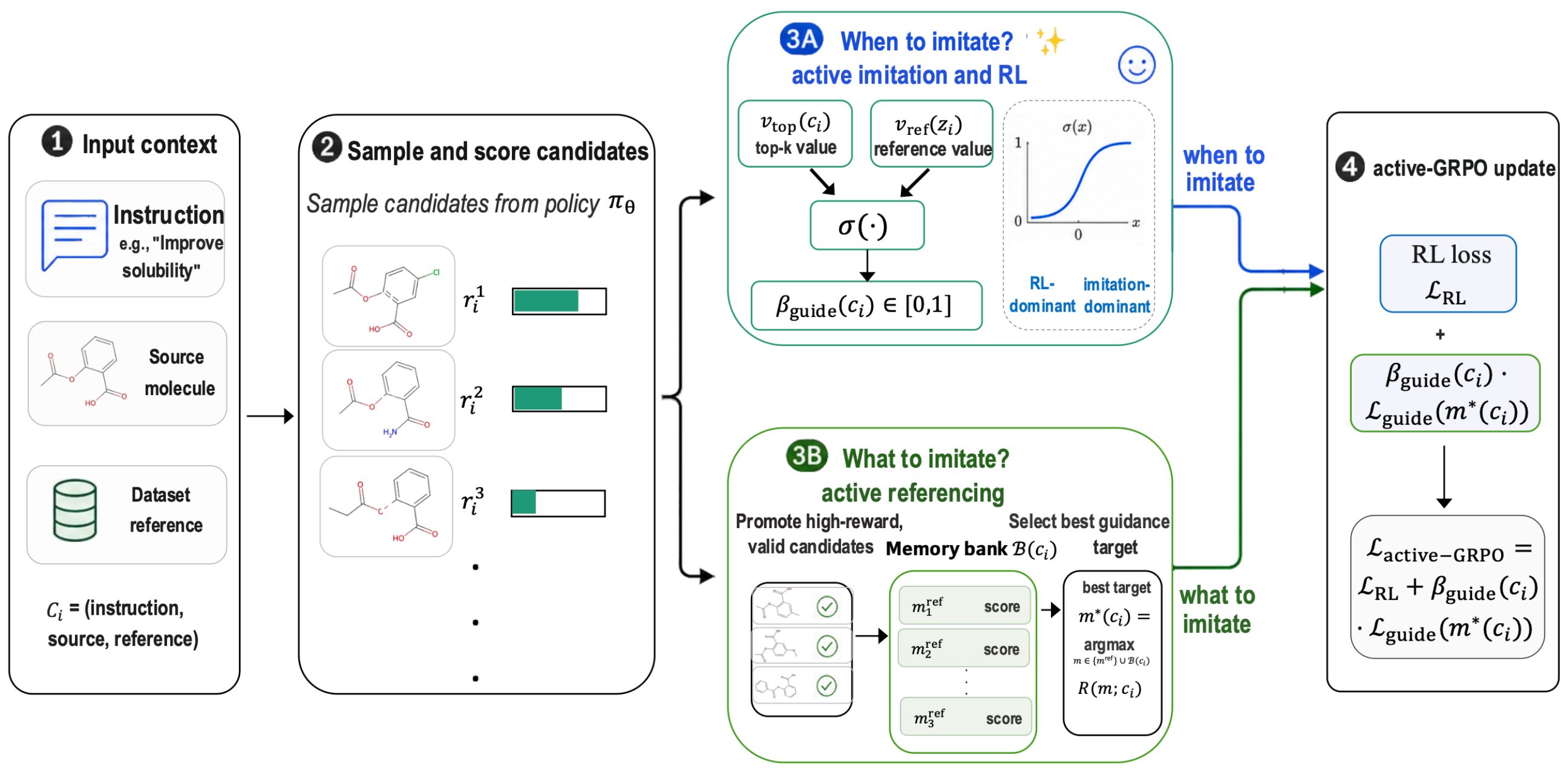}
    \caption{Overview of \algname. The method augments reference-guided policy optimization with two active decisions: when to imitate, through a dynamic guidance weight, and what to imitate, through active reference selection over promoted policy-discovered candidates.}
    \label{fig:method_overview}
    \vspace{-0.3cm}
\end{figure*}

\paragraph{Active Imitate-Reinforce.}
We adaptively blend imitation and reinforcement learning to achieve robust policy improvement, allowing the learner to switch between imitating an oracle and improving via RL based on online relative performance. In our setting, the dataset reference serves as the initial oracle, and policy-generated candidates can themselves become improved oracles once they surpass it. Rather than enforcing a hard switch, we implement this idea as a smooth, per-instance guidance weight.
For each context $c_i$, let
\[
v_{\mathrm{top}}(c_i)
=
\frac{1}{k}\sum_{j\in \mathrm{Top}\text{-}k} R(\hat m_{i,j};c_i)
\]
be the mean reward of the top-$k$ sampled candidates from the current rollout group. Setting $k=1$ recovers the best sampled candidate; larger $k$ yields a smoother top-$k$ average. We define the context-dependent guidance weight
\[
\beta_{\mathrm{guide}}(c_i)
=
\beta_{\min}
+
(\beta_{\max}-\beta_{\min})
\,\sigma\!\left(
-\alpha\bigl(v_{\mathrm{top}}(c_i)-v_{\mathrm{ref}}(z_i)\bigr)
\right),
\]
where $\sigma(\cdot)$ is the logistic sigmoid and $\alpha > 0$ controls sharpness of the transition. The semantics are:
\begin{itemize}[noitemsep, wide,labelwidth=0pt, labelindent=0pt]
    \item {Policy lags the reference} ($v_{\mathrm{top}} < v_{\mathrm{ref}}$): $\beta_{\mathrm{guide}}$ is large, and the update is imitation-dominant.
    \item {Policy surpasses the reference} ($v_{\mathrm{top}} > v_{\mathrm{ref}}$): $\beta_{\mathrm{guide}}$ shrinks, and the update shifts toward reinforcement learning.
\end{itemize}
The bounds $\beta_{\min}$ and $\beta_{\max}$ control, respectively, the residual imitation pressure once the policy has surpassed the reference and the maximum imitation weight when it lags. The choice of $\beta_{\min}$ in particular determines the late-training regime: $\beta_{\min} = 0$ recovers pure RL in the limit, while $\beta_{\min} > 0$ retains a residual self-distillation signal toward the current best target Appendix~\ref{app:betaregime}.

\paragraph{\textbf{{Active referencing.}}}
Active referencing replaces the static dataset reference with the best policy-discovered candidate available so far. For each context $c_i$, Active-GRPO maintains a capacity-limited memory bank
\[
\mathcal{B}(c_i)=\{(m_\ell,r_\ell)\}_{\ell=1}^{|\mathcal{B}(c_i)|},
\qquad
|\mathcal{B}(c_i)|\le K,
\]
{initialized with the dataset-provided reference molecule,
}
\[
{
\mathcal{B}(c_i)\leftarrow
\{(m_{\mathrm{ref},i}, v_{\mathrm{ref}}(z_i))\},
}
\]
{and is subsequently augmented with policy-generated molecules that are promoted during training. Thus, each bank entry consists of a candidate guidance molecule $m_\ell$ and its reward $r_\ell=R(m_\ell;c_i)$ under the same conditioning context.}
The bank is keyed by a deterministic per-example identifier and persists across training steps and epochs, thus each example accumulates a best-so-far set over the course of training.
{We store up to $K$ candidates rather than only the current best one so that the bank represents a small per-instance candidate set. Although our main implementation selects the maximum-reward candidate as the active target, the same bank can support more robust reference statistics, such as top-$k$ averaging or diversity-aware target selection.}

\textit{Promotion and eviction.} A generated molecule $\widehat{m}$ is promoted into the bank when it improves on the current reference reward by a margin $\delta$ and satisfies a task-specific admissibility predicate:
\[
\mathrm{Promote}(\hat m\mid c_i)
=
\mathbf{1}\!\Big[
R(\hat m;c_i)>v_{\mathrm{ref}}(z_i)+\delta
\;\wedge\;
Q(\hat m;c_i)
\;\wedge\;
\mathrm{Valid}(\hat m)
\Big].
\]
The margin $\delta$ guards against promoting near-tie noise; the predicate $Q(\widehat{m}; c_i)$ enforces task-specific hard constraints (in our experiments, a minimum Tanimoto similarity to the input molecule) that are kept separate from the scalar reward to prevent structurally invalid candidates from entering the bank merely because they score high on one reward component. When the bank is full, the lowest-reward entry is evicted. Promoted molecules are canonicalized before insertion so that supervision targets a deterministic SMILES surface form (Appendix~\ref{app:bank},~\ref{app:canonicalization}).

\textit{Active guidance target.}The target supplied to $\mathcal{L}_{\mathrm{guide}}$ is the highest-reward entry currently in the bank,
\[
m^*(c_i)
=
\arg\max_{(m,r)\in \mathcal{B}(c_i)} r.
\]
Thus, once the policy discovers a molecule better than the original dataset reference, subsequent guidance distills from the best available candidate rather than from the static reference.

\textit{Optimization Objective.}
\label{sec:objective}
For a minibatch of size $B$, \algname optimizes
\[
\mathcal{L}_{\algname}
=
\mathcal{L}_{\mathrm{RL}}
+
\frac{1}{B}\sum_{i=1}^B
\beta_{\mathrm{guide}}(c_i)\,
\mathcal{L}_{\mathrm{guide}}^{(i)}\!\bigl(m^*(c_i)\bigr).
\]
Crucially, the guidance loss is weighted \emph{per instance} rather than by a batch-averaged coefficient. This preserves the intended adaptive behavior: different examples within the same minibatch may simultaneously operate in different regimes of imitation versus self-improvement, depending on how their current top-$k$ rollouts compare to their respective references.

Together, the two mechanisms make reference guidance robust by construction across the spectrum of reference quality. Relative to RePO~\citep{li2026repo}, \algname introduces two essential changes: a dynamic, context-dependent guidance coefficient that decides when and how strongly to imitate, and an active referencing mechanism that replaces static reference guidance with the best available target discovered during training. Algorithm~\ref{alg:apiar} in Appendix~\ref{app:training_algorithm} summarizes the complete training procedure; additional implementation, synchronization, reproducibility, and hyperparameter details are provided in Appendix~\ref{app:implementation}.

%% file: experiments.tex
\section{Experiments}
\label{sec:experiments}

We evaluate \algname on instruction-conditioned molecular optimization benchmarks and compare it against RePO~\citep{li2026repo} and related baselines. Our experiments address four questions: 
\begin{itemize}[noitemsep, wide,labelwidth=0pt, labelindent=0pt]
    \item Q1. Does Active-GRPO improve molecular optimization performance over fixed-reference guidance? (Section~\ref{sec:main_results})
    \item Q2. Are the gains explained by simpler alternatives such as iterative self-distillation or stronger fixed references? (Sections~\ref{sec:main_results}; ablations in Sections~\ref{sec:ablation})
    \item Q3. Does Active-GRPO's advantage grow with the optimization headroom of each instance, as the static-reference ceiling argument predicts? (Section~\ref{sec:headroom})
    \item Q4. Do training-time dynamics match the intended adaptive mechanism?  (Section~\ref{sec:dynamics})
\end{itemize}

\subsection{Experimental Setup}
\label{sec:exp_setup}

\paragraph{Benchmarks and metrics.}
We evaluate on TOMG-Bench~\citep{li2024tomg}, focusing on the \textsc{MolOpt} subtasks LogP, MR, and QED. These tasks provide a controlled setting for reference-guided, instruction-conditioned molecular optimization. Following RePO~\citep{li2026repo}, we report Success Rate (SR), Tanimoto Similarity (Sim) with Morgan fingerprints, and the composite metric SR$\times$Sim, which summarizes the trade-off between task success and structural preservation. Additional evaluations on TOMG-Bench \textsc{MolEdit}, hard-example splits, and longer-horizon settings are reported in Appendix~\ref{app:additional_experiments}.

\paragraph{Baselines and variants.}
We compare against zero-shot inference, GRPO-only training without answer-level guidance, RePO as the fixed-reference baseline, and two alternatives that test simpler explanations: Iterative SFT, which captures self-distillation without active policy improvement, and Offline-strengthened RePO, which replaces the dataset reference with a stronger fixed target before training. We also include two ablations of \algname: $\beta$-only, which uses active imitate-reinforce without active referencing, and bank-only, which uses active referencing with a fixed guidance weight.

\paragraph{Training and reporting protocol.}
All trained methods share the same backbone, reward family, rollout budget, decoding rule, and evaluation pipeline under a matched 40GB-A100 configuration, which may differ from prior RePO reports~\citep{li2026repo}; we therefore interpret results as matched relative comparisons. We report mean $\pm$ standard error over three seeds (zero-shot once), with significance assessed via per-example paired bootstrap (10{,}000 resamples). Implementation details, hyperparameters, and timing are in Appendix~\ref{app:implementation} and~\ref{app:wallclock}.

\subsection{Main Results on TOMG-Bench MolOpt}
\label{sec:main_results}

\begin{table*}[t]
\centering
\caption{Main results on TOMG-Bench \textsc{MolOpt}. We report SR$\times$Sim, the standard composite metric balancing task success and structural preservation. All trained methods are reported as mean $\pm$ standard error over three seeds; zero-shot is evaluated once. \algname achieves the best performance on all three subtasks and the highest average score. All trained methods are run under the same matched 40GB-A100 configuration.}
\label{tab:molopt_main}
\vspace{-0.2cm}
\resizebox{\textwidth}{!}{%
\begin{tabular}{lcccc}
\toprule
Method & LogP SR$\times$Sim $\uparrow$ & MR SR$\times$Sim $\uparrow$ & QED SR$\times$Sim $\uparrow$ & Avg SR$\times$Sim $\uparrow$ \\
\midrule
Zero-shot & 0.1700 & 0.1314 & 0.1114 & 0.1376 \\
GRPO-only & 0.1222 $\pm$ 0.0112 & 0.0956 $\pm$ 0.0132 & 0.0699 $\pm$ 0.0061 & 0.0959 \\
Iterative SFT & 0.1888 $\pm$ 0.0266 & 0.1736 $\pm$ 0.0200 & 0.1257 $\pm$ 0.0116 & 0.1627 \\
Offline-strengthened RePO & 0.1974 $\pm$ 0.0076 & 0.1853 $\pm$ 0.0045 & 0.1130 $\pm$ 0.0108 & 0.1652 \\
RePO & 0.1877 $\pm$ 0.0076 & 0.1860 $\pm$ 0.0089 & 0.1258 $\pm$ 0.0033 & 0.1665 \\
\midrule
\textbf{\algname\ (ours)} & \textbf{0.1977 $\pm$ 0.0104} & \textbf{0.1904 $\pm$ 0.0100} & \textbf{0.1440 $\pm$ 0.0081} & \textbf{0.1773} \\
\bottomrule
\end{tabular}%
}
\vspace{-0.2cm}
\end{table*}

Table~\ref{tab:molopt_main} reports the main results on TOMG-Bench \textsc{MolOpt}. 
\algname obtains the highest SR$\times$Sim on all three subtasks and the best overall average. 
Relative to RePO, the gains are +0.0148 on LogP, +0.0090 on MR, and +0.0206 on QED, all significant under paired bootstrap testing ($p<0.001$; Appendix~\ref{app:significance}).
The comparison rules out two simpler explanations. Iterative SFT improves over GRPO-only but stays below both RePO and \algname, showing that self-distillation alone is insufficient. Offline-strengthened RePO is the strongest non-adaptive alternative, yet still falls below \algname, indicating the gain is not explained by replacing the reference with a stronger fixed target. The advantage comes from making guidance adaptive and per-instance, not merely stronger.
\vspace{-0.2cm}
\paragraph{Success--similarity trade-off.}
\algname's improvement comes primarily from higher success rates. Averaged across the three subtasks, SR rises from 0.2017 (RePO) to 0.2249, while average similarity decreases from 0.8281 to 0.7906. This is consistent with the design: active imitate-reinforce reduces unnecessary imitation pressure, and active referencing lets the policy move beyond the local neighborhood of the fixed reference when doing so improves reward. As we show in Section~\ref{sec:qualitative}, much of RePO's higher similarity reflects no-op behavior --- copying the source molecule --- rather than successful structural preservation under a real edit. Full per-metric results are in Appendix~\ref{app:molopt_breakdown}.

\vspace{-0.2cm}
\subsection{Ablation: Active Imitate-Reinforce and Active Referencing Are Complementary}
\label{sec:ablation}

Table~\ref{tab:ablation_results} isolates the two core mechanisms in \algname. On average SR$\times$Sim, neither component alone improves over RePO: the active-imitate-reinforce-only variant underperforms RePO, while the active-referencing-only variant is approximately tied. The full method, however, improves over all variants by a clear margin, indicating that active imitate-reinforce and active referencing are complementary rather than redundant.

\begin{table*}[t]
\centering
\caption{Ablation results on TOMG-Bench \textsc{MolOpt}. Neither active imitate-reinforce nor active referencing alone explains the gain. The full method achieves the best score, indicating that the two mechanisms play complementary roles and are most effective when coupled.}
\label{tab:ablation_results}
\vspace{-0.2cm}
\resizebox{\textwidth}{!}{%
\begin{tabular}{lccccc}
\toprule
Variant & LogP SR$\times$Sim $\uparrow$ & MR SR$\times$Sim $\uparrow$ & QED SR$\times$Sim $\uparrow$ & Avg SR$\times$Sim $\uparrow$ & $\Delta$ vs Full \\
\midrule
RePO & 0.1877 $\pm$ 0.0076 & 0.1860 $\pm$ 0.0089 & 0.1258 $\pm$ 0.0033 & 0.1665 & $-0.0108$ \\
\algname\ (active-imitate-reinforce only) & 0.1861 $\pm$ 0.0099 & 0.1791 $\pm$ 0.0101 & 0.1220 $\pm$ 0.0098 & 0.1624 & $-0.0149$ \\
\algname\ (active-referencing only) & 0.1920 $\pm$ 0.0078 & 0.1781 $\pm$ 0.0046 & 0.1275 $\pm$ 0.0072 & 0.1659 & $-0.0114$ \\
\textbf{\algname\ (full)} & \textbf{0.1977 $\pm$ 0.0104} & \textbf{0.1904 $\pm$ 0.0100} & \textbf{0.1440 $\pm$ 0.0081} & \textbf{0.1773} & --- \\
\bottomrule
\end{tabular}%
}
\vspace{-0.2cm}
\end{table*}

This pattern supports the design. Active imitate-reinforce controls \emph{when} and how strongly the model should imitate, while active referencing controls \emph{what} target to imitate. Using only one mechanism leaves coordination incomplete: active-imitate-reinforce only can reduce imitation pressure but still imitates a fixed target, whereas active-referencing only can update the target but cannot adapt guidance strength. The full method combines both, allowing the policy to reduce stale-reference pressure while distilling from stronger policy-discovered targets.

\vspace{-0.2cm}
\subsection{Optimization-Headroom Conditional Analysis}
\label{sec:headroom}

The static-reference ceiling argument predicts a specific empirical pattern: \algname should help most on instances where the static source-reference offers the weakest guidance toward the requested edit --- that is, where there is more optimization headroom beyond the reference. We test this directly.
For each test example, we compute an optimization-headroom score from the source molecule's original property value: larger original LogP/MR for the decrease tasks, and $1 - \text{QED}$ for increase-QED. We partition test examples into quintiles and compare Active-GRPO against RePO within each bin. (This measures optimization headroom rather than an independent reference-quality gap; here the source molecule also serves as the static anchor.)

The results support the central mechanism of \algname. The trend is clearest for LogP, where the \algname–RePO gain in SR×Sim grows monotonically from +0.002 in Q1 to +0.031 in Q5. MR and QED show positive but noisier patterns, with QED gaining in both low- and high-headroom regimes. \algname is most useful precisely where the static reference is least informative: as headroom grows, active imitate-reinforce reduces unnecessary imitation pressure, and active referencing supplies stronger policy-discovered targets. Full per-subtask tables are in Appendix~\ref{app:headroom_full}.

\begin{figure*}[t]
\centering

\begin{minipage}[t]{0.60\textwidth}
    \vspace{0pt}
    \hspace*{-0.03\textwidth}%
    \includegraphics[width=0.74\linewidth]{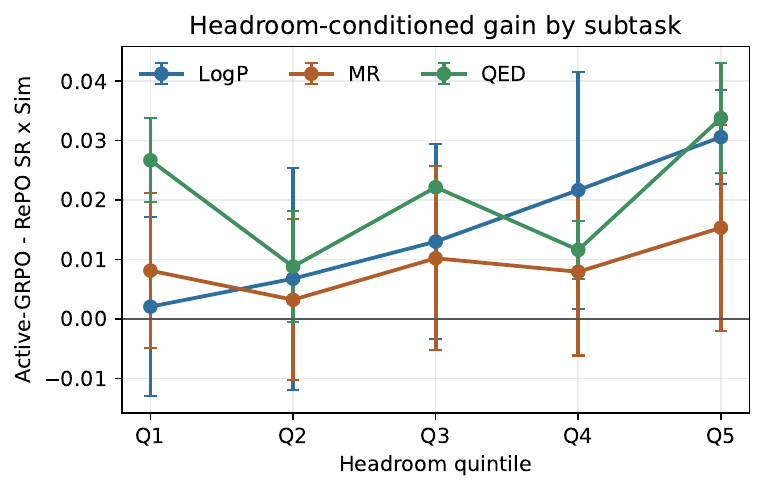}
\end{minipage}
\hspace{-0.005\textwidth}
\begin{minipage}[t]{0.35\textwidth}
    \vspace{1.0em}
    \centering
    {\bfseries\small LogP headroom breakdown}
    \vspace{0.4em}

    \footnotesize
    \setlength{\tabcolsep}{3pt}
    \begin{tabular}{lccc}
    \toprule
    Bin & Range & $\Delta$SR & $\Delta$SR$\times$Sim \\
    \midrule
    Q1 & $(-3.91, 1.30]$ & +0.007 & +0.002 \\
    Q2 & $(1.30, 2.25]$  & +0.014 & +0.007 \\
    Q3 & $(2.25, 2.94]$  & +0.024 & +0.013 \\
    Q4 & $(2.94, 3.63]$  & +0.032 & +0.022 \\
    Q5 & $(3.63, 7.36]$  & +0.045 & +0.031 \\
    \bottomrule
    \end{tabular}
\end{minipage}

\vspace{-0.4em}
\caption{Optimization-headroom conditional analysis. Left: \algname--RePO gain in SR$\times$Sim across headroom quintiles for all three \textsc{MolOpt} subtasks. Right: the LogP numerical breakdown, where the gain increases monotonically from the lowest-headroom bin to the highest-headroom bin. \fix{Full per-subtask numerical tables are reported in Appendix~\ref{app:headroom_full}.}}
\label{fig:headroom}
\vspace{-0.5cm}
\end{figure*}

\subsection{Training Dynamics Match the Intended Mechanism}
\label{sec:dynamics}
\vspace{-0.2cm}

\begin{figure*}[t]
    \centering
    \includegraphics[width=\textwidth]{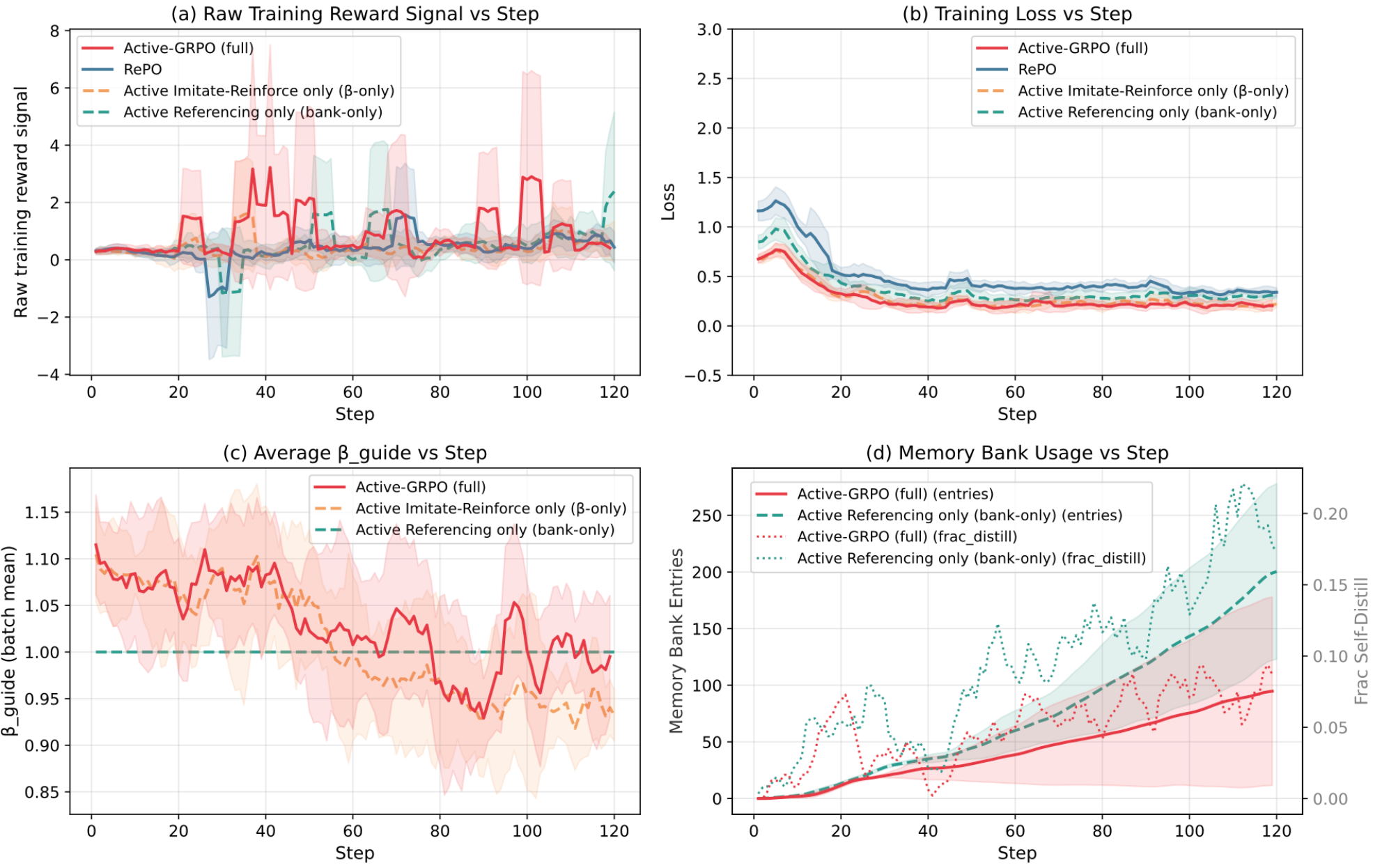}
    \caption{\algname training dynamics over three seeds, with mean $\pm$ standard deviation. Panel (a) shows the raw training reward signal logged by the trainer, rather than evaluation SR$\times$Sim; panels (b)--(d) show training loss, average guidance weight, and memory-bank/self-distillation activity.}
    \label{fig:training_dynamics}
\vspace{0.3cm}
\end{figure*}

We verify that \algname behaves as designed by inspecting training-time statistics. Figure~\ref{fig:training_dynamics} reports raw training reward, training loss, average guidance weight, and memory-bank usage. Three patterns are consistent with the intended adaptive mechanism:
\vspace{-0.2cm}
\begin{itemize}[noitemsep, wide,labelwidth=0pt, labelindent=0pt]
    \item \textit{Guidance shifts from imitation toward RL.} The average guidance weight $\beta_{\mathrm{guide}}$ starts around 1.1 --- above the midpoint, indicating early-training imitation dominance --- and decreases to about 0.93, indicating the policy has caught up to or surpassed references on a substantial fraction of examples.
    \item \textit{The memory bank fills steadily.} \algname accumulates 160–190 promoted entries by the end of training, confirming the policy regularly produces molecules superior to their dataset references.
    \item \textit{Self-distillation activates meaningfully.} Roughly 17–22\% of examples receive guidance from a policy-promoted target rather than the dataset reference, confirming that active referencing is a substantive contributor to training.
\end{itemize}
\vspace{-0.2cm}
Together, these dynamics show that \algname is not merely reweighting the RePO loss: it actively changes both the strength and the target of guidance during training, as predicted by the active reasoning paradigm.

\subsection{Qualitative Case Studies} 
\label{sec:qualitative}
\vspace{-0.2cm}

We inspect representative examples in which \algname succeeds and RePO fails. Across LogP, MR, and QED, two characteristic failure modes of fixed-reference guidance recur: RePO often (i) returns the source molecule essentially unchanged or (ii) makes a structural modification that moves the property in the worse direction. \algname instead tends to make targeted edits that satisfy the property objective while preserving most of the input structure. For example, \algname expands a ring to increase LogP, shortens a ring system to decrease MR, and introduces a small heteroatom change to decrease QED.

These patterns help explain \algname's success–similarity trade-off: much of RePO's higher average similarity comes from no-op outputs that satisfy the similarity term because no edit was made. We quantify this directly via a no-op failure rate — the fraction of outputs with $\text{Sim}(\widehat{m}, m_{\mathrm{src}}) \ge 0.98$ that nevertheless fail the property objective. Across all three subtasks, \algname reduces no-op failures relative to RePO (Appendix~\ref{app:case_studies}, Figure~\ref{fig:noop_failure_rate}), confirming that its slightly lower similarity reflects more frequent successful editing rather than excessive structural deviation. 

\subsection{Additional Evaluations}
\label{sec:additional_eval}

We report additional evaluations in Appendix~\ref{app:additional_experiments}, including qualitative case studies and visible reasoning-trace examples (Appendix~\ref{app:case_studies}), hyperparameter sensitivity (Appendix~\ref{app:sensitivity}), hard-example stress tests (Appendix~\ref{app:hard_data}), matched \textsc{MolEdit} structural optimization (Appendix~\ref{app:moledit}), and longer-horizon single-seed evaluation (Appendix~\ref{app:fullscale}). These results support the main finding while clarifying the method's scope. On a ZINC-derived hard subset, \algname maintains a small but consistent advantage over RePO. On \textsc{MolEdit}, \algname improves average SR$\times$Sim over RePO and performs best on AddComponent and SubComponent, suggesting that the adaptive mechanisms transfer beyond property-only optimization. The longer-horizon evaluation further shows competitive average performance under shared-policy training. Broader discovery-leaning and strongly multi-objective settings remain important directions for future work.

%% file: conclusion.tex
\section{Conclusion}
\label{sec:conclusion}

We introduced \algname, an active extension of reference-guided policy optimization that adapts both \emph{when} to imitate through a context-dependent guidance weight and \emph{what} to imitate through active referencing over policy-discovered candidates. On TOMG-Bench \textsc{MolOpt}, \algname achieves the best average SR$\times$Sim and improves over RePO on all three subtasks under matched multi-seed evaluation, with ablations showing that dynamic guidance and active referencing play complementary roles. Our results suggest that fixed-reference guidance is most limiting when the source molecule leaves substantial optimization headroom, while adaptive guidance better supports continued policy improvement. A current limitation is that our evaluation remains centered on reference-guided molecular optimization; extending the framework to broader discovery settings with weak or absent references may require stronger candidate proposal and exploration mechanisms. One promising direction is to combine \algname with active sampling or curriculum construction, focusing training compute on prompts where the policy, reference, and reward signal disagree most.

%% file: acknowledgement.tex
\subsubsection*{Acknowledgements}
We thank Xuan Li, Bo Han for their helpful discussion. This work is supported by Donald and Delia Baxter Foundation Faculty Scholar award, the Weintz family foundation and AI4Biomedicine fund.

%% file: supp_implementation.tex
\section{Implementation and Reproducibility Details}
\label{app:implementation}

This appendix provides implementation details omitted from the main algorithmic presentation, including the precise RL objective, rollout synchronization, memory bank semantics, canonicalization of guidance targets, and hyperparameter settings.

\subsection{RL Objective and KL Regularization}
\label{app:kl}

The RL term in \algname follows the same GRPO-style objective used in our RePO baseline. For a minibatch of prompt groups, we compute group-relative advantages
\[
\hat A_{i,j}
=
\frac{r_{i,j}-\bar r_i}{\sigma_{r,i}+\varepsilon},
\qquad
\bar r_i=\frac{1}{G}\sum_{j=1}^G r_{i,j},
\]
and optimize
\[
\mathcal{L}_{\mathrm{RL}}
=
-\frac{1}{\sum_{i,j,t} m^{\mathrm{cmp}}_{i,j,t}}
\sum_{i,j,t}
\Big[
\rho_{i,j,t}\hat A_{i,j}
-
\beta_{\mathrm{KL}}D^{\mathrm{KL}}_{i,j,t}
\Big]m^{\mathrm{cmp}}_{i,j,t}.
\]

Here $\rho_{i,j,t}$ denotes the usual token-level policy ratio term used in GRPO-style updates, and $D^{\mathrm{KL}}_{i,j,t}$ is the KL regularizer to a frozen reference model $\pi_{\mathrm{ref}}$.

\paragraph{KL estimator.}
In our implementation, we use the same KL estimator as in the RePO/GRPO training stack used for all baselines and variants. This choice is held fixed across RePO and \algname so that all comparisons differ only in the adaptive guidance and active reference mechanisms. %

\subsection{Rollout Policy, Reference Model, and Synchronization}
\label{app:sync}

Rollouts are generated from a rollout policy corresponding to the current training policy at the beginning of each rollout phase. This rollout policy is kept fixed while computing rewards, advantages, and losses for the resulting minibatch, and is then refreshed for the next rollout phase after the model update. This is the standard lagged-policy setup used in policy optimization.

The KL reference model $\pi_{\mathrm{ref}}$ is frozen throughout training. It is not periodically refreshed, and serves only as a stabilizing reference for KL regularization. All methods in our comparisons use the same reference-model treatment.

In the actual implementation, rollout generation is performed through the same inference backend for RePO and \algname, with identical decoding settings except where explicitly varied in ablations.

\subsection{Memory Bank Semantics and Persistence}
\label{app:bank}

For each training instance, \algname maintains a memory bank
\[
\mathcal{B}(c_i)=\{(m_\ell,r_\ell)\}_{\ell=1}^{|\mathcal{B}(c_i)|},
\qquad
|\mathcal{B}(c_i)|\le K.
\]
The bank persists across training steps and across epochs: when the same training example reappears, the previously accumulated bank is reused rather than reinitialized. Algorithmically, this means \algname performs persistent best-so-far self-distillation rather than purely local within-step imitation.

\paragraph{Keying semantics.}
The memory bank is indexed by a deterministic identifier of the training instance. In our experiments, this effectively means one bank per example, rather than matching free-form instructions by textual similarity. This avoids ambiguity arising from paraphrased instructions and makes the active reference mechanism well-defined across repeated visits to the same example.

\paragraph{Promotion and eviction.}
A generated molecule $\hat m$ is promoted when it exceeds the current reference reward by margin $\delta$ and satisfies the admissibility predicate $Q(\hat m;c_i)$. If the bank is already full, the lowest-reward entry is evicted. This yields a capacity-limited best-so-far memory bank for each example.

\paragraph{Age-based eviction.}
Unless otherwise stated, the main experiments use persistent capacity-based storage only: entries are removed only through reward-based eviction when the bank is full. %

\subsection{Guidance Targets, Canonicalization, and Answer Masking}
\label{app:canonicalization}

The active guidance target for instance $i$ is
\[
m_i^*
=
\arg\max_{m\in \{m_{\mathrm{ref},i}\}\cup \mathcal{B}(c_i)} R(m;c_i).
\]
All molecules inserted into the memory bank are canonicalized before storage, and the selected active target is canonicalized again before constructing the teacher-forced answer sequence. This ensures that the guidance loss is applied to a deterministic SMILES representation for each promoted molecule.

Answer-level supervision is implemented by masking only the token span corresponding to the final answer molecule:
\[
\mathcal{L}_{\mathrm{guide}}^{(i)}(m)
=
-
\frac{
\sum_t \log \pi_\theta(m_t\mid x_i,m_{<t})\,m^{\mathrm{ans}}_{i,t}
}{
\sum_t m^{\mathrm{ans}}_{i,t}+\varepsilon
}.
\]
Thus, the reasoning trace is preserved as model-generated context, while supervision is applied only to the answer molecule itself.

\subsection{Task-Specific Admissibility Predicate \texorpdfstring{$Q(\hat m;c)$}{Q(m; c)}}
\label{app:qfilter}

The admissibility predicate $Q(\hat m;c)$ is separated from the scalar reward for conceptual and practical reasons. The reward $R(\hat m;c)$ ranks candidates among admissible outputs, whereas $Q(\hat m;c)$ enforces task-specific hard constraints that define whether a candidate is eligible for promotion into the memory bank.

In the molecule editing experiments, $Q(\hat m;c)$ is instantiated as a hard minimum-similarity constraint to the input molecule, computed using Tanimoto similarity between Morgan fingerprints. The purpose of this separation is to prevent structurally invalid or semantically off-task candidates from entering the bank merely because they achieve a high scalar reward under one component of $R$. In other task settings, $Q(\hat m;c)$ may instead encode scaffold preservation, substructure constraints, synthesizability filters, or other admissibility conditions.

\subsection{Adaptive Guidance Regimes}
\label{app:betaregime}

The adaptive guidance coefficient is
\[
\beta_{\mathrm{guide}}(c_i)
=
\beta_{\min}
+
(\beta_{\max}-\beta_{\min})
\,\sigma\!\left(
-\alpha\bigl(v_{\mathrm{top}}(c_i)-v_{\mathrm{ref}}(z_i)\bigr)
\right).
\]
The value of $\beta_{\min}$ determines the asymptotic behavior of the algorithm.

If $\beta_{\min}=0$, then once the policy reliably outperforms the current reference, the algorithm transitions to pure RL in the limit. If $\beta_{\min}>0$, then \algname retains a residual self-distillation signal toward the current best-so-far target, even in late training. We keep this parameter fixed within each experiment and report its value in the hyperparameter table below.

\subsection{Sampling and Training Hyperparameters}
\label{app:hyperparams}

Unless otherwise noted, all methods compared in the main text share the same backbone model, optimizer, rollout budget, reward function, and initialization. \algname differs from RePO only in the addition of the adaptive guidance coefficient and the active reference update mechanism.

The main hyperparameters introduced by \algname are:
\begin{itemize}
    \item $k$: number of top samples used in $v_{\mathrm{top}}(c)$;
    \item $\alpha$: sharpness of the sigmoid switching rule;
    \item $\beta_{\min},\beta_{\max}$: lower and upper bounds for the guidance coefficient;
    \item $\delta$: promotion margin over the current reference reward;
    \item $K$: memory-bank capacity per training instance.
\end{itemize}

For reproducibility, we report the exact values used in the main experiments in Table~\ref{tab:apiar-hparams}.

\begin{table}[t]
\centering
\caption{Main hyperparameters used in the matched \algname experiments.}
\label{tab:apiar-hparams}
\small
\setlength{\tabcolsep}{6pt}
\begin{tabular}{lc}
\toprule
Hyperparameter & Value \\
\midrule
\multicolumn{2}{l}{\textit{Shared sampling/training parameters}} \\
Number of sampled outputs $G$ & 3 \\
KL coefficient $\beta_{\mathrm{KL}}$ & 0.04 \\
Sampling temperature (base / high) & 0.9 / 1.3 \\
Maximum completion length & 1024 \\
\midrule
\multicolumn{2}{l}{\textit{\algname-specific parameters}} \\
Top-$k$ size for $v_{\mathrm{top}}$ & $k=\lceil 0.33G\rceil=1$ \\
Guidance lower bound $\beta_{\min}$ & 0.3 \\
Guidance upper bound $\beta_{\max}$ & 1.5 \\
Sigmoid sharpness $\alpha$ & 3.0 \\
Promotion margin $\delta$ & 0.05 \\
Memory-bank capacity $K$ & 5 \\
Similarity threshold in $Q$ & 0.2 \\
\bottomrule
\end{tabular}
\end{table}

\subsection{Training Algorithm}
\label{app:training_algorithm}

\begin{algorithm}[t]
\caption{\algname Training}
\label{alg:apiar}
\begin{algorithmic}[1]
\Require Dataset $\mathcal{D}=\{(c_i,m_{\mathrm{ref},i})\}_{i=1}^N$, system prompt $p_{\mathrm{sys}}$, policy $\pi_\theta$, frozen reference model $\pi_{\mathrm{ref}}$
\For{each training step}
    \State Sample minibatch $\{(c_i,m_{\mathrm{ref},i})\}_{i=1}^B$ from $\mathcal{D}$
    \For{$i=1,\dots,B$}
        \State If $\mathcal{B}(c_i)$ is uninitialized, set $\mathcal{B}(c_i)\gets\{(m_{\mathrm{ref},i}, v_{\mathrm{ref}}(z_i))\}$
        \State Construct prompt $x_i \gets [p_{\mathrm{sys}};c_i]$
        \State Sample $G$ outputs $\{o_{i,j}\}_{j=1}^G \sim \pi_\theta(\cdot\mid x_i)$
        \State Extract candidate molecules $\{\hat m_{i,j}\}_{j=1}^G$
        \State Compute rewards $r_{i,j}\gets R(\hat m_{i,j};c_i)$
        \State Compute group-relative advantages $\hat A_{i,j}$
        \State Compute $v_{\mathrm{top}}(c_i)$ from the top-$k$ rewards
        \State Compute
        \[
        \beta_{\mathrm{guide}}(c_i)\gets
        \beta_{\min}+(\beta_{\max}-\beta_{\min})
        \sigma\!\left(-\alpha\bigl(v_{\mathrm{top}}(c_i)-v_{\mathrm{ref}}(z_i)\bigr)\right)
        \]
        \For{$j=1,\dots,G$}
            \If{$\mathrm{Promote}(\hat m_{i,j}\mid c_i)=1$}
                \State Insert $(\mathrm{canon}(\hat m_{i,j}), r_{i,j})$ into $\mathcal{B}(c_i)$, evicting the lowest-reward entry if necessary
            \EndIf
        \EndFor
        \State Set $m_i^*$ to the molecule with highest stored reward in $\mathcal{B}(c_i)$
        \State Compute answer mask $m^{\mathrm{ans}}_{i,t}$ and guidance loss $\mathcal{L}_{\mathrm{guide}}^{(i)}(m_i^*)$
    \EndFor
    \State Compute minibatch RL loss $\mathcal{L}_{\mathrm{RL}}$
    \State Compute total loss
    \[
    \mathcal{L}_{\algname}
    \gets
    \mathcal{L}_{\mathrm{RL}}
    +
    \frac{1}{B}\sum_{i=1}^B
    \beta_{\mathrm{guide}}(c_i)\,
    \mathcal{L}_{\mathrm{guide}}^{(i)}(m_i^*)
    \]
    \State Update $\theta$ by gradient descent on $\mathcal{L}_{\algname}$
\EndFor
\end{algorithmic}
\end{algorithm}

%% file: related_works.tex
\section{Related Work}\label{sec:related}
 
\paragraph{LLM Reasoning via SFT and Reinforcement Learning.}
Eliciting structured, multi-step reasoning from large language models has become a central research thread. Early work showed that simple prompting strategies such as chain-of-thought \citep{wei2022chain, kojima2022large} can substantially improve performance on tasks requiring deliberation. Building on this, supervised fine-tuning on curated reasoning traces \citep{zelikman2022star, yu2023metamath, muennighoff2025s1} further internalizes step-by-step problem solving, while reinforcement learning with verifiable rewards (RLVR)—using programmatic checkers as the reward signal—has driven recent advances in mathematical and code reasoning \citep{guo2025deepseek, lambert2024tulu, shao2024deepseekmath}. Group Relative Policy Optimization (GRPO) \citep{shao2024deepseekmath} has in particular become a default RLVR algorithm, replacing PPO's value network with group-relative advantage estimation. However, both paradigms exhibit characteristic failure modes that motivate our work: answer-only SFT can suppress intermediate reasoning and harm generalization \citep{chu2025sft, lin2023unlocking}, while RLVR struggles with sparse feedback in domains where most samples fail verification gates \citep{andrychowicz2017hindsight, riedmiller2018learning}—a pathology that is especially acute in instruction-based molecular optimization, where tight similarity and validity constraints make zero-reward batches the norm rather than the exception. \algname inherits the RLVR formulation but addresses the sparse-reward pathology through reference guidance that adapts on a per-instance basis to the policy's current capability.
 
\paragraph{LLMs for Scientific Discovery and Molecular Optimization.}
A growing body of work applies LLMs to scientific reasoning, ranging from domain-specialized pretraining \citep{taylor2022galactica} and surveys of LLM-driven discovery \citep{ai4science2023impact} to autonomous experimental agents in chemistry \citep{boiko2023autonomous} and drug discovery \citep{stokes2020deep}. Within this landscape, \emph{instruction-based molecular optimization} has emerged as a uniquely demanding testbed: given a source molecule and a natural-language instruction, the model must propose a structurally similar yet property-improved candidate \citep{jin2018learning, jin2020hierarchical, edwards2022translation, li2024empowering}. The TOMG-Bench suite \citep{li2024tomg} formalized this task across editing, optimization, and customized-generation subtasks, and subsequent work has explored prompting \citep{li2024empowering}, surrogate-model integration, and tool-augmented agents. A separate line of research evaluates molecular generation against multi-property oracles using benchmarks such as GuacaMol \citep{brown2019guacamol}, MOSES \citep{polykovskiy2020molecular}, the Therapeutics Data Commons \citep{huang2021therapeutics}, and PMO \citep{gao2022sample}. Unlike these oracle-driven settings, our work targets the instruction-conditional regime in which reference molecules are provided alongside the instruction, similarity constraints are tight, and the central challenge is \emph{leveraging}—rather than simply imitating—the supplied references. This regime is precisely where the static-reference ceiling we identify becomes consequential: a fixed reference is simultaneously the most natural form of guidance and the most direct source of brittleness when its quality varies across instances.
 
\paragraph{Reference-Guided and Demonstration-Augmented Policy Optimization.}
Anchoring reinforcement learning to expert demonstrations has a long history. Behavior cloning provides a pure imitation baseline, while methods that combine demonstrations with RL—DAPG \citep{rajeswaran2017learning}, DQfD \citep{hester2018deep}, demonstration-augmented PPO \citep{nair2018overcoming}, and advantage-weighted regression \citep{peng2019advantage}—use demonstrations to densify reward signal and stabilize exploration in sparse-reward settings, while DAgger \citep{ross2011reduction} addresses distribution shift in pure imitation. In the LLM setting, Reference-guided Policy Optimization (RePO) \citep{li2026repo}—our most direct prior work—adapts the demonstration-augmented RL idea to instruction-based molecular optimization by combining a GRPO-style RL term with a reference-guidance term that fixes the reasoning trajectory and supervises only the final answer. A common assumption underlies all of these methods: demonstrations are treated as \emph{fixed targets}, with the policy implicitly assumed never to surpass them. This assumption is the source of the static-reference ceiling we characterize in this paper: when references are weak, noisy, or misaligned with the instruction \citep{gao2024impact, belkhale2023data}, the imitation signal actively pulls the policy away from better solutions it would otherwise discover. The closest conceptual precedent for our approach is robust policy improvement \citep{liu2023blending}, which adaptively blends imitation and RL based on online relative performance against a fixed oracle. \algname departs from this entire family in two key respects, both motivated by our active-reasoning paradigm: it makes the choice of \emph{when} to imitate per-instance and policy-conditional, and—uniquely—it makes \emph{what} to imitate adaptive by replacing the reference itself once the policy surpasses it.
 
\paragraph{Self-Improvement and Iterative Refinement in LLMs.}
A complementary line of work trains LLMs to improve themselves by generating, filtering, and re-training on their own outputs. Expert Iteration \citep{anthony2017thinking} formalized this loop in the RL-with-tree-search setting; STaR \citep{zelikman2022star} adapted it to reasoning by bootstrapping on self-generated rationales filtered by answer correctness. Reinforced Self-Training (ReST) \citep{gulcehre2023reinforced} and ReST-EM \citep{singh2023beyond} alternate between sampling, reward-based filtering, and supervised fine-tuning on filtered outputs. RAFT \citep{dong2023raft} similarly fine-tunes on top-ranked samples under a reward model, and Self-Rewarding Language Models \citep{yuan2024self} let the model serve as its own reward source. These methods share a key insight: the policy's own best outputs can serve as a stronger training target than fixed external data, provided the filtering signal is reliable. \algname builds on this insight but contributes a distinct mechanism that bridges this line of work with the reference-guided one above. Rather than \emph{discarding} external references in favor of self-generated ones—or \emph{preserving} them as fixed targets—\algname \emph{integrates} the two through an explicit per-instance comparison: when the dataset reference still outperforms the policy's own samples, it imitates the reference; when its samples surpass the reference, it self-improves and promotes its own discovery into the imitation target. This is the operational form of our active-reasoning paradigm: the policy actively decides \emph{when} to imitate versus reinforce, and continuously upgrades \emph{what} it imitates. To our knowledge, \algname is the first method to apply adaptive, policy-conditional reference replacement to instruction-based molecular optimization, where tight similarity constraints and competing property objectives make the choice between imitation and self-improvement particularly consequential.

%% file: supp_experiment.tex
\section{Additional Experimental Details}
\label{app:additional_experiments}

This appendix provides additional experimental details and extended results for Section~\ref{sec:experiments}. We include the full per-metric breakdown, statistical tests, headroom-conditioned analyses for all \textsc{MolOpt} subtasks, computational overhead, qualitative case studies, hyperparameter sensitivity, hard-example stress tests, and longer-horizon single-seed evaluations.

\subsection{Full Per-Metric Results on TOMG-Bench MolOpt}
\label{app:molopt_breakdown}

Table~\ref{tab:molopt_per_metric} reports the full per-metric breakdown for the main TOMG-Bench \textsc{MolOpt} results. The main text emphasizes SR$\times$Sim as the standard composite metric, while this table separates the success and similarity terms to expose the optimization--preservation trade-off.

\begin{table*}[t]
\centering
\caption{Full per-metric results on TOMG-Bench \textsc{MolOpt}. Trained methods are reported as mean $\pm$ standard error over three seeds; zero-shot is evaluated once.}
\label{tab:molopt_per_metric}
\resizebox{\textwidth}{!}{%
\begin{tabular}{lcccccc}
\toprule
Method & LogP SR $\uparrow$ & LogP Sim $\uparrow$ & MR SR $\uparrow$ & MR Sim $\uparrow$ & QED SR $\uparrow$ & QED Sim $\uparrow$ \\
\midrule
Zero-shot
& 0.2308 & 0.7368 & 0.1666 & 0.7885 & 0.1396 & 0.7980 \\
GRPO-only
& 0.1405 $\pm$ 0.0150 & 0.8728 $\pm$ 0.0149
& 0.1069 $\pm$ 0.0163 & 0.8974 $\pm$ 0.0152
& 0.0779 $\pm$ 0.0079 & 0.8997 $\pm$ 0.0143 \\
Iterative SFT
& 0.2509 $\pm$ 0.0318 & 0.7493 $\pm$ 0.0127
& 0.2114 $\pm$ 0.0244 & 0.8213 $\pm$ 0.0030
& 0.1539 $\pm$ 0.0130 & 0.8158 $\pm$ 0.0090 \\
Offline-strengthened RePO
& 0.2526 $\pm$ 0.0018 & 0.7814 $\pm$ 0.0291
& 0.2206 $\pm$ 0.0072 & 0.8412 $\pm$ 0.0250
& 0.1336 $\pm$ 0.0172 & 0.8524 $\pm$ 0.0268 \\
RePO
& 0.2368 $\pm$ 0.0096 & 0.7925 $\pm$ 0.0023
& 0.2181 $\pm$ 0.0120 & 0.8534 $\pm$ 0.0058
& 0.1501 $\pm$ 0.0044 & 0.8383 $\pm$ 0.0030 \\
\textbf{\algname}
& \textbf{0.2613 $\pm$ 0.0139} & 0.7565 $\pm$ 0.0126
& \textbf{0.2314 $\pm$ 0.0098} & 0.8221 $\pm$ 0.0104
& \textbf{0.1820 $\pm$ 0.0134} & 0.7932 $\pm$ 0.0184 \\
\bottomrule
\end{tabular}%
}
\end{table*}

Table~\ref{tab:molopt_per_metric} shows that \algname's gain comes primarily from higher success rates. RePO and GRPO-only often preserve higher similarity, but this similarity can reflect conservative or no-op behavior rather than successful optimization. This motivates the qualitative analysis in Section~\ref{sec:qualitative}.

\subsection{Statistical Significance}
\label{app:significance}

We assess statistical significance using a paired bootstrap over per-example SR$\times$Sim differences. For each method and subtask, we compute SR$\times$Sim for every evaluation example in each seed, pool the matched per-example observations across the three seeds, and bootstrap paired indices with 10{,}000 resamples. For each resample, we compute the mean difference between \algname and the comparison method. The reported one-sided $p$-value is the fraction of bootstrap resamples for which this mean difference is non-positive. Table~\ref{tab:significance} reports the resulting SR$\times$Sim differences and $p$-values.

\begin{table*}[t]
\centering
\caption{Paired-bootstrap significance tests for SR$\times$Sim. Differences are \algname minus the comparison method. Bootstrap resampling is performed over matched per-example observations pooled across the three seeds.}
\label{tab:significance}
\resizebox{\textwidth}{!}{%
\begin{tabular}{lcccccc}
\toprule
Comparison & LogP $\Delta$ & LogP $p$ & MR $\Delta$ & MR $p$ & QED $\Delta$ & QED $p$ \\
\midrule
\algname\ vs RePO
& +0.0148 & $<0.001$
& +0.0090 & $<0.001$
& +0.0206 & $<0.001$ \\
\algname\ vs Offline-strengthened RePO
& +0.0115 & $<0.001$
& +0.0126 & $<0.001$
& +0.0357 & $<0.001$ \\
\algname\ vs Iterative SFT
& +0.0105 & $<0.001$
& +0.0168 & $<0.001$
& +0.0204 & $<0.001$ \\
\algname\ vs GRPO-only
& +0.0895 & $<0.001$
& +0.0953 & $<0.001$
& +0.0792 & $<0.001$ \\
\algname\ vs $\beta$-only
& +0.0169 & $<0.001$
& +0.0142 & $<0.001$
& +0.0235 & $<0.001$ \\
\algname\ vs bank-only
& +0.0113 & $<0.001$
& +0.0134 & $<0.001$
& +0.0188 & $<0.001$ \\
\bottomrule
\end{tabular}%
}
\end{table*}

\subsection{Headroom-Conditional Analysis on All Subtasks}
\label{app:headroom_full}

In Section~\ref{sec:headroom}, we report the headroom-conditioned analysis in the main text, with the LogP breakdown shown alongside the curve. Here we provide the complete protocol and the corresponding extended analysis for all three \textsc{MolOpt} subtasks. Tables~\ref{tab:headroom_logp_full}, \ref{tab:headroom_mr_full}, and \ref{tab:headroom_qed_full} report the per-bin results for LogP, MR, and QED, respectively.

For each test example, we compute an optimization-headroom score from the source molecule's original property value. For decrease-LogP and decrease-MR tasks, larger original LogP/MR values indicate more room for improvement. For increase-QED, we define headroom as \(1-\mathrm{QED}\). We then partition test examples into quintiles and report the \algname--RePO gap within each bin. These bins measure optimization headroom rather than an independent reference-quality gap; in this benchmark, the source molecule is the static anchor against which optimization is requested.

\begin{table*}[t]
\centering
\caption{Headroom-conditioned analysis on LogP. The \algname--RePO gap increases with optimization headroom.}
\label{tab:headroom_logp_full}
\resizebox{\textwidth}{!}{%
\begin{tabular}{lccccc}
\toprule
Quintile & Gap range & $\Delta$ SR & APIAR SR$\times$Sim & RePO SR$\times$Sim & $\Delta$ SR$\times$Sim \\
\midrule
Q1 & $(-3.91, 1.30]$ & +0.007 & 0.168 & 0.167 & +0.002 \\
Q2 & $(1.30, 2.25]$ & +0.014 & 0.173 & 0.160 & +0.007 \\
Q3 & $(2.25, 2.94]$ & +0.024 & 0.175 & 0.159 & +0.013 \\
Q4 & $(2.94, 3.63]$ & +0.032 & 0.183 & 0.161 & +0.022 \\
Q5 & $(3.63, 7.36]$ & +0.045 & 0.189 & 0.157 & +0.031 \\
\bottomrule
\end{tabular}%
}
\end{table*}

\begin{table*}[t]
\centering
\caption{Headroom-conditioned analysis on MR. We bucket test examples by the original MR value of the source molecule; larger values indicate more room for improvement under the decrease-MR objective.}
\label{tab:headroom_mr_full}
\resizebox{\textwidth}{!}{%
\begin{tabular}{lccccc}
\toprule
Quintile & Gap range & $\Delta$ SR & APIAR SR$\times$Sim & RePO SR$\times$Sim & $\Delta$ SR$\times$Sim \\
\midrule
Q1 & $(19.97, 74.51]$   & +0.013 & 0.137 & 0.129 & +0.008 \\
Q2 & $(74.51, 84.89]$   & +0.003 & 0.152 & 0.149 & +0.003 \\
Q3 & $(84.89, 93.59]$   & +0.014 & 0.158 & 0.148 & +0.010 \\
Q4 & $(93.59, 102.92]$  & +0.014 & 0.164 & 0.156 & +0.008 \\
Q5 & $(102.92, 142.84]$ & +0.022 & 0.176 & 0.161 & +0.015 \\
\bottomrule
\end{tabular}%
}
\end{table*}

\begin{table*}[t]
\centering
\caption{Headroom-conditioned analysis on QED. We bucket test examples by $1-\mathrm{QED}$ of the source molecule; larger values indicate more room for improvement under the increase-QED objective.}
\label{tab:headroom_qed_full}
\resizebox{\textwidth}{!}{%
\begin{tabular}{lccccc}
\toprule
Quintile & Gap range & $\Delta$ SR & APIAR SR$\times$Sim & RePO SR$\times$Sim & $\Delta$ SR$\times$Sim \\
\midrule
Q1 & $(0.05, 0.15]$ & +0.041 & 0.129 & 0.102 & +0.027 \\
Q2 & $(0.15, 0.20]$ & +0.016 & 0.118 & 0.109 & +0.009 \\
Q3 & $(0.20, 0.27]$ & +0.035 & 0.129 & 0.107 & +0.022 \\
Q4 & $(0.27, 0.38]$ & +0.020 & 0.113 & 0.102 & +0.012 \\
Q5 & $(0.38, 0.82]$ & +0.047 & 0.130 & 0.096 & +0.034 \\
\bottomrule
\end{tabular}%
}
\end{table*}

\fix{Because the test split does not provide a separate gold reference molecule for each example, we do not construct a separate reference-quality stratification; source-property headroom is the relevant task-level proxy for how informative the static source-reference is for the requested edit.}

\subsection{Computational Overhead}
\label{app:wallclock}

Table~\ref{tab:wallclock} reports matched wall-clock measurements for RePO and \algname. Both methods use the same model architecture and rollout setup; the additional computations in \algname, including dynamic guidance weighting and memory-bank operations, are lightweight relative to online generation and model optimization.

\begin{table}[t]
\centering
\caption{Wall-clock comparison between RePO and \algname. Timing is measured over matched 30-step runs on 4$\times$A100-SXM4-40GB with DeepSpeed ZeRO-3.}
\label{tab:wallclock}
\small
\setlength{\tabcolsep}{5pt}
\begin{tabular}{lcccc}
\toprule
Method & Total time & Sec./step & Samples/sec. & Overhead \\
\midrule
RePO & 4720.8s & 157.4 & 0.61 & baseline \\
\algname & 4657.5s & 155.3 & 0.62 & $-1.3\%$ \\
\bottomrule
\end{tabular}
\end{table}

\subsection{Qualitative Case Studies}
\label{app:case_studies}

\begin{figure*}[t]
    \centering
    \includegraphics[width=\textwidth]{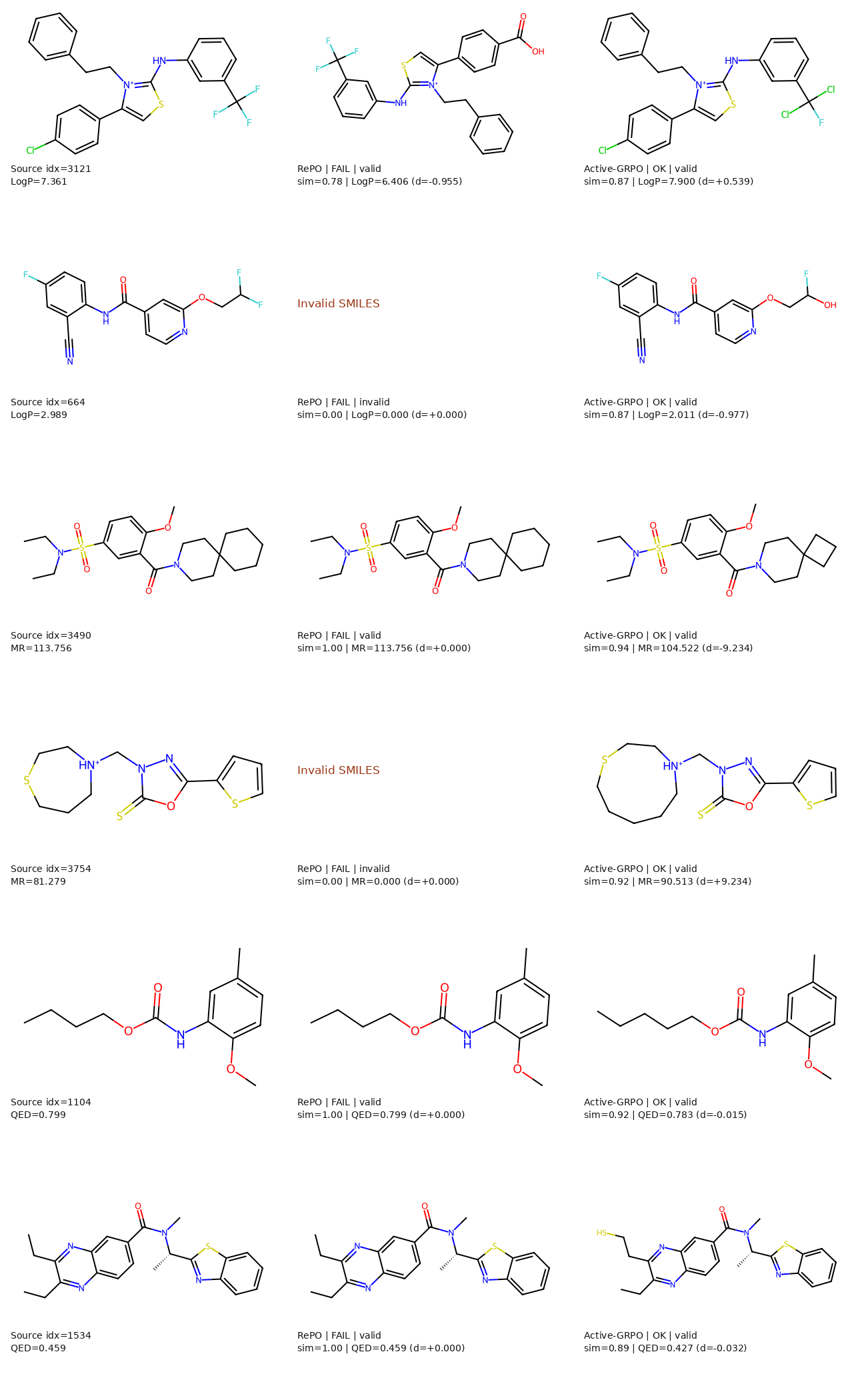}
    \caption{Qualitative case studies across LogP, MR, and QED. 
    \algname makes targeted structural edits that satisfy the requested property objective while preserving much of the input structure. 
    RePO often either copies the source molecule, yielding high similarity but no property improvement, or produces an invalid or wrong-direction edit.}
    \label{fig:case_studies_appendix}
\end{figure*}

Figure~\ref{fig:case_studies_appendix} shows representative examples where \algname succeeds and RePO fails across LogP, MR, and QED tasks. 
Each row contains the source molecule, the RePO output, and the \algname output, together with similarity and property-change annotations.

\begin{figure}[t]
    \centering
    \includegraphics[width=0.7\columnwidth]{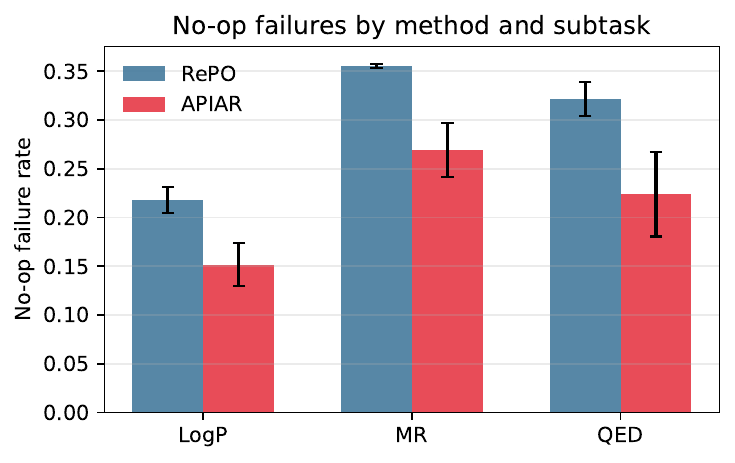}
    \caption{No-op failure rate by method and subtask. A no-op failure is defined as an output with high similarity to the source molecule, $\mathrm{Sim}(\hat m,m_{\mathrm{src}})\ge 0.98$, that nevertheless fails the requested property objective. Lower is better.}
    \label{fig:noop_failure_rate}
\end{figure}

\paragraph{No-op failure analysis.}
To quantify the copy-like failure mode observed in the qualitative examples, we measure the no-op failure rate:
\[
\mathrm{NoOpFail}
=
\mathbf{1}\{\mathrm{Sim}(\hat m,m_{\mathrm{src}})\ge 0.98
\;\wedge\;
\mathrm{Success}=0\}.
\]
Figure~\ref{fig:noop_failure_rate} reports this rate by method and subtask. Across \textsc{MolOpt} subtasks, \algname reduces no-op failures relative to RePO, supporting the qualitative observation that active imitation and RL with active referencing encourages more effective edits rather than simply preserving the source molecule.

\paragraph{Visible reasoning traces.}
We further inspect visible rationales emitted before the final SMILES answer. These are model-generated text outputs, not hidden chain-of-thought. The examples below show a recurring pattern: RePO often states a plausible chemical strategy, but the final extracted SMILES is unchanged, invalid, or inconsistent with the stated objective. In contrast, \algname more often converts the stated optimization direction into a valid targeted edit.

\begin{tcolorbox}[tracebox,title={Example 1: Increasing LogP via hydrophobic substitution}]
\small
\textbf{Instruction.}
Optimize \smiles{FC(F)(F)c1cccc(Nc2scc(-c3ccc(Cl)cc3)[n+]2CCc2ccccc2)c1} to have a higher LogP value.

\vspace{2pt}
\setlength{\tabcolsep}{2pt}
\begin{tabular}{>{\raggedright\arraybackslash}p{0.12\linewidth} p{0.37\linewidth} p{0.37\linewidth}}
\toprule
 & \textbf{Active-GRPO} & \textbf{RePO} \\
\midrule
Output &
\smiles{FC(Cl)(Cl)c1cccc(...)c1} &
\smiles{FC(F)(F)c1cccc(...C(=O)[OH]...)c1} \\
Eval. &
Success, sim $=0.873$ &
Failure, sim $=0.780$ \\
Visible rationale &
``replace the fluorine atoms with chlorine atoms'' to increase hydrophobicity &
states that a larger hydrophobic substituent should be introduced, but returns a carboxylic-acid-containing edit \\
\bottomrule
\end{tabular}

\vspace{2pt}
\textbf{Interpretation.}
Active-GRPO's rationale and edit are directionally aligned with the target property.
RePO identifies the need for higher hydrophobicity, but the generated molecule introduces a polar acid group and fails the objective.
\end{tcolorbox}

\begin{tcolorbox}[tracebox,title={Example 2: Decreasing LogP by adding polarity}]
\small
\textbf{Instruction.}
Modify \smiles{N\#Cc1cc(F)ccc1NC(=O)c1ccnc(OCC(F)F)c1} to decrease its LogP value.

\vspace{2pt}
\setlength{\tabcolsep}{2pt}
\begin{tabular}{>{\raggedright\arraybackslash}p{0.12\linewidth} p{0.37\linewidth} p{0.37\linewidth}}
\toprule
 & \textbf{Active-GRPO} & \textbf{RePO} \\
\midrule
Output &
\smiles{N\#Cc1cc(F)ccc1NC(=O)c1ccnc(OCC(F)O)c1} &
\smiles{\textbackslash{}(N\#Cc1cc(F)ccc1NC(=O)c1ccnc(OCC(F)F)c1\textbackslash{})} \\
Eval. &
Success, sim $=0.865$ &
Invalid, sim $=0.000$ \\
Visible rationale &
introduces a hydroxyl group to increase hydrophilicity and reduce LogP &
describes removing/replacing fluorine atoms, but the final extracted answer retains formatting delimiters and is invalid \\
\bottomrule
\end{tabular}

\vspace{2pt}
\textbf{Interpretation.}
Active-GRPO maps the property direction to a valid polar substitution.
RePO's rationale is broadly plausible, but the final answer is not a valid molecule.
\end{tcolorbox}

\begin{tcolorbox}[tracebox,title={Example 3: Decreasing MR by simplifying a ring system}]
\small
\textbf{Instruction.}
Modify \smiles{CCN(CC)S(=O)(=O)c1ccc(OC)c(C(=O)N2CCC3(CCCCC3)CC2)c1} to decrease its MR value.

\vspace{2pt}
\setlength{\tabcolsep}{2pt}
\begin{tabular}{>{\raggedright\arraybackslash}p{0.12\linewidth} p{0.37\linewidth} p{0.37\linewidth}}
\toprule
 & \textbf{Active-GRPO} & \textbf{RePO} \\
\midrule
Output &
\smiles{CCN(CC)S(=O)(=O)c1ccc(OC)c(C(=O)N2CCC3(CCC3)CC2)c1} &
\smiles{CCN(CC)S(=O)(=O)c1ccc(OC)c(C(=O)N2CCC3(CCCCC3)CC2)c1} \\
Eval. &
Success, sim $=0.940$ &
Failure, sim $=1.000$ \\
Visible rationale &
reduces the bulky fused ring component to lower MR &
states that the ring should be simplified, but returns the original molecule \\
\bottomrule
\end{tabular}

\vspace{2pt}
\textbf{Interpretation.}
This example illustrates a no-op failure.
RePO's natural-language strategy is reasonable, but the final SMILES is unchanged.
Active-GRPO makes a conservative structural simplification that satisfies the target.
\end{tcolorbox}

\begin{tcolorbox}[tracebox,title={Example 4: Lowering QED while avoiding a no-op}]
\small
\textbf{Instruction.}
Optimize \smiles{CCc1nc2ccc(C(=O)N(C)[C@@H](C)c3nc4ccccc4s3)cc2nc1CC} to have a lower QED value.

\vspace{2pt}
\setlength{\tabcolsep}{2pt}
\begin{tabular}{>{\raggedright\arraybackslash}p{0.12\linewidth} p{0.37\linewidth} p{0.37\linewidth}}
\toprule
 & \textbf{Active-GRPO} & \textbf{RePO} \\
\midrule
Output &
\smiles{CCc1nc2ccc(C(=O)N(C)[C@@H](C)c3nc4ccccc4s3)cc2nc1CCS} &
\smiles{CCc1nc2ccc(C(=O)N(C)[C@@H](C)c3nc4ccccc4s3)cc2nc1CC} \\
Eval. &
Success, sim $=0.889$ &
Failure, sim $=1.000$ \\
Visible rationale &
proposes a sulfur-containing modification and returns a non-trivial edit &
describes simplifying the molecule, but returns the original SMILES \\
\bottomrule
\end{tabular}

\vspace{2pt}
\textbf{Interpretation.}
Active-GRPO preserves high similarity while changing the molecule enough to alter the property.
RePO again produces a no-op despite describing an intended edit.
\end{tcolorbox}

Overall, these examples suggest that Active-GRPO improves not merely by changing output distributions, but by making the final molecular action more consistent with the model's stated property-level reasoning.
This qualitative pattern matches the quantitative no-op analysis: RePO more frequently emits unchanged or invalid molecules on examples where a small targeted edit is required.

\subsection{Hyperparameter Sensitivity}
\label{app:sensitivity}

We evaluate the sensitivity of \algname to three key hyperparameters: the sigmoid sharpness $\alpha$, the memory-bank capacity $K$, and the promotion margin $\delta$. We use MolOpt-LogP as the test bed and vary one parameter at a time while holding the others at their default values: $\alpha=3.0$, $K=5$, and $\delta=0.05$. Table~\ref{tab:sensitivity} reports the single-seed sweep results.

\begin{table}[t]
\centering
\caption{Hyperparameter sensitivity on MolOpt-LogP. We vary one parameter at a time around the default setting $\alpha=3.0$, $K=5$, $\delta=0.05$.}
\label{tab:sensitivity}
\small
\setlength{\tabcolsep}{6pt}
\begin{tabular}{lccc}
\toprule
Configuration & SR $\uparrow$ & Sim $\uparrow$ & SR$\times$Sim $\uparrow$ \\
\midrule
$\alpha=1.0$ & 0.2350 & 0.7911 & 0.1859 \\
$\alpha=3.0$ (default) & 0.2613 & 0.7565 & 0.1977 \\
$\alpha=5.0$ & 0.2390 & 0.8004 & 0.1913 \\
$\alpha=10.0$ & 0.2476 & 0.7851 & 0.1944 \\
\midrule
$K=1$ & 0.2336 & 0.8207 & 0.1917 \\
$K=3$ & 0.2112 & 0.8194 & 0.1731 \\
$K=5$ (default) & 0.2613 & 0.7565 & 0.1977 \\
$K=10$ & 0.2372 & 0.7974 & 0.1891 \\
\midrule
$\delta=0.01$ & 0.2522 & 0.8050 & \textbf{0.2030} \\
$\delta=0.05$ (default) & 0.2613 & 0.7565 & 0.1977 \\
$\delta=0.10$ & 0.2244 & 0.8198 & 0.1840 \\
$\delta=0.20$ & 0.2534 & 0.8012 & \textbf{0.2030} \\
\bottomrule
\end{tabular}
\end{table}

The sweep suggests that \algname is not highly sensitive to a single carefully tuned hyperparameter. Across all configurations, SR$\times$Sim remains in the range 0.1731--0.2030. The default setting is near the top of the sweep, while several nearby settings, especially alternative promotion margins, match or slightly exceed the default. The lowest score occurs at $K=3$, but even this setting remains close to the main RePO baseline, indicating that the method is reasonably robust across plausible hyperparameter choices.

\subsection{Hard-Example Stress Test}
\label{app:hard_data}

We further evaluate RePO and \algname on a hard-example subset derived from ZINC, containing molecules whose initial properties leave substantially larger room for optimization. Table~\ref{tab:hard_data} reports results on this subset. This stress test probes whether active imitation and RL with active referencing remains beneficial when successful edits are rarer and the optimization problem is more difficult.

\begin{table}[t]
\centering
\caption{Hard-example stress test on a ZINC-derived difficult subset. Results are mean $\pm$ standard error over three seeds.}
\label{tab:hard_data}
\resizebox{\columnwidth}{!}{%
\begin{tabular}{llccc}
\toprule
Subtask & Method & SR $\uparrow$ & Sim $\uparrow$ & SR$\times$Sim $\uparrow$ \\
\midrule
LogP & RePO & 0.1011 $\pm$ 0.0097 & \textbf{0.9207 $\pm$ 0.0037} & 0.0930 $\pm$ 0.0085 \\
LogP & \algname & \textbf{0.1204 $\pm$ 0.0134} & 0.8859 $\pm$ 0.0242 & \textbf{0.1061 $\pm$ 0.0089} \\
\midrule
MR & RePO & 0.0757 $\pm$ 0.0091 & \textbf{0.9479 $\pm$ 0.0061} & 0.0716 $\pm$ 0.0081 \\
MR & \algname & \textbf{0.0807 $\pm$ 0.0115} & 0.9257 $\pm$ 0.0209 & \textbf{0.0742 $\pm$ 0.0089} \\
\midrule
QED & RePO & 0.0468 $\pm$ 0.0069 & \textbf{0.9538 $\pm$ 0.0058} & 0.0446 $\pm$ 0.0063 \\
QED & \algname & \textbf{0.0545 $\pm$ 0.0142} & 0.9299 $\pm$ 0.0329 & \textbf{0.0504 $\pm$ 0.0113} \\
\bottomrule
\end{tabular}%
}
\end{table}

On this hard subset, both methods become more conservative: success rates are substantially lower than on the standard \textsc{MolOpt} split, while similarities remain high. Nevertheless, \algname improves SR and SR$\times$Sim over RePO on all three subtasks, suggesting that adaptive guidance remains useful when the optimization problem is more difficult.

\subsection{Longer-Horizon Single-Seed Evaluation}
\label{app:fullscale}

We additionally evaluate RePO and \algname in a longer-horizon single-seed setting using Qwen2.5-3B-Instruct on A100-80GB GPUs. We train for 4 epochs on OpenMolIns-light, with 1500 training examples per task group, and evaluate on TOMG-Bench using Success Rate (SR), Similarity (Sim), and SR$\times$Sim. Table~\ref{tab:fullscale_tomgbench} reports the complete results. To evaluate transfer across related objectives, we train one shared policy for the property-based objectives (LogP, MR, and QED) and one shared policy for the structure-based objectives (AddComponent, DelComponent, and SubComponent), rather than training a separate policy for each objective. We report this evaluation as complementary evidence to the main multi-seed comparison.

\begin{table*}[t]
\centering
\caption{Longer-horizon single-seed evaluation on TOMG-Bench. We train one shared policy for structure-based objectives and one shared policy for property-based objectives. We report Success Rate (SR), Similarity (Sim), and SR$\times$Sim; higher is better.}
\label{tab:fullscale_tomgbench}
\small
\setlength{\tabcolsep}{6pt}
\begin{tabular}{lllccc}
\toprule
Task type & Objective & Metric & Zero-shot & RePO & \algname \\
\midrule
\multirow{10}{*}{\makecell[l]{Structure-based\\optimization}}
& \multirow{3}{*}{AddComponent}
& SR & 0.123 & 0.408 & \textbf{0.458} \\
& & Sim & 0.573 & \textbf{0.722} & 0.718 \\
& & SR$\times$Sim & 0.071 & 0.295 & \textbf{0.329} \\
\cmidrule(lr){2-6}
& \multirow{3}{*}{DelComponent}
& SR & 0.250 & 0.339 & \textbf{0.389} \\
& & Sim & 0.601 & 0.752 & \textbf{0.754} \\
& & SR$\times$Sim & 0.150 & 0.255 & \textbf{0.293} \\
\cmidrule(lr){2-6}
& \multirow{3}{*}{SubComponent}
& SR & 0.151 & 0.502 & \textbf{0.600} \\
& & Sim & 0.657 & 0.752 & \textbf{0.760} \\
& & SR$\times$Sim & 0.099 & 0.377 & \textbf{0.456} \\
\cmidrule(lr){2-6}
& Avg & SR$\times$Sim & 0.107 & 0.309 & \textbf{0.359} \\
\midrule
\multirow{10}{*}{\makecell[l]{Property-based\\optimization}}
& \multirow{3}{*}{LogP}
& SR & 0.309 & 0.443 & \textbf{0.669} \\
& & Sim & 0.628 & \textbf{0.711} & 0.609 \\
& & SR$\times$Sim & 0.194 & 0.315 & \textbf{0.408} \\
\cmidrule(lr){2-6}
& \multirow{3}{*}{MR}
& SR & 0.249 & 0.505 & \textbf{0.574} \\
& & Sim & 0.630 & \textbf{0.709} & 0.600 \\
& & SR$\times$Sim & 0.157 & \textbf{0.358} & 0.345 \\
\cmidrule(lr){2-6}
& \multirow{3}{*}{QED}
& SR & 0.222 & \textbf{0.348} & 0.330 \\
& & Sim & 0.613 & \textbf{0.722} & 0.635 \\
& & SR$\times$Sim & 0.136 & \textbf{0.251} & 0.210 \\
\cmidrule(lr){2-6}
& Avg & SR$\times$Sim & 0.162 & 0.308 & \textbf{0.321} \\
\bottomrule
\end{tabular}
\end{table*}

The longer-horizon evaluation also reveals a limitation: under the shared property-policy setting, \algname underperforms RePO on QED in SR$\times$Sim. Since LogP, MR, and QED are optimized by a single shared policy in this experiment, per-objective trade-offs may differ from those obtained under separate per-property training. At the same time, \algname achieves the best average SR$\times$Sim across the three property objectives in this longer-horizon setting. Because this evaluation is single-seed, we use it as complementary evidence about transfer across related objectives, while relying on the main multi-seed \textsc{MolOpt} comparison for the primary statistical claim.

\subsection{Matched MolEdit Structural Optimization}
\label{app:moledit}

We further evaluate RePO and \algname on TOMG-Bench \textsc{MolEdit}, which tests structure-based molecular editing through AddComponent, DelComponent, and SubComponent objectives. Both methods are trained under the same matched 4$\times$A100-40GB configuration for three seeds, using the same backbone, rollout budget, reward functions, decoding setup, and evaluation pipeline. We train one shared structure-editing policy over the three \textsc{MolEdit} objectives and evaluate each subtask separately. Table~\ref{tab:moledit} reports the full SR, Sim, and SR$\times$Sim breakdown.

\begin{table*}[t]
\centering
\caption{Matched 3-seed \textsc{MolEdit} structural-optimization results. We train one shared structure-editing policy over AddComponent, DelComponent, and SubComponent, and report mean $\pm$ standard error over three seeds.}
\label{tab:moledit}
\small
\setlength{\tabcolsep}{4pt}
\begin{tabular}{llccc}
\toprule
Subtask & Method & SR $\uparrow$ & Sim $\uparrow$ & SR$\times$Sim $\uparrow$ \\
\midrule
\multirow{2}{*}{AddComponent}
& RePO & 0.2003 $\pm$ 0.0176 & \textbf{0.7579 $\pm$ 0.0054} & 0.1517 $\pm$ 0.0125 \\
& \algname & \textbf{0.2282 $\pm$ 0.0104} & 0.7515 $\pm$ 0.0128 & \textbf{0.1716 $\pm$ 0.0098} \\
\midrule
\multirow{2}{*}{DelComponent}
& RePO & \textbf{0.1393 $\pm$ 0.0064} & 0.8661 $\pm$ 0.0106 & \textbf{0.1205 $\pm$ 0.0041} \\
& \algname & 0.1142 $\pm$ 0.0184 & \textbf{0.8927 $\pm$ 0.0219} & 0.1012 $\pm$ 0.0137 \\
\midrule
\multirow{2}{*}{SubComponent}
& RePO & 0.3067 $\pm$ 0.0046 & 0.7901 $\pm$ 0.0027 & 0.2423 $\pm$ 0.0042 \\
& \algname & \textbf{0.3245 $\pm$ 0.0151} & \textbf{0.8056 $\pm$ 0.0102} & \textbf{0.2615 $\pm$ 0.0139} \\
\midrule
Avg & RePO & 0.2154 & 0.8047 & 0.1715 \\
Avg & \algname & \textbf{0.2223} & \textbf{0.8166} & \textbf{0.1781} \\
\bottomrule
\end{tabular}
\end{table*}